\documentclass[journal]{IEEEtran}

\usepackage[utf8]{inputenc}
\usepackage{tabu}
\usepackage[pdftex]{graphicx}
\usepackage{amsmath}
\usepackage{amssymb}
\usepackage{array}
\usepackage[caption=false,font=footnotesize]{subfig}
\usepackage{makecell}
\usepackage{flushend}
\usepackage{cite}
\begin{document}
%\title{Enhancing Looming Selectivity during UAV Agile Flights by Incorporating a Spatial-temporal Distributed Presynaptic Connections into an LGMD Model
%}
\title{Enhancing LGMD's Looming Selectivity\\ for UAV with Spatial-temporal Distributed Presynaptic Connections
}
\author{Jiannan Zhao, Hongxin Wang, Nicola Bellotto, Cheng Hu, Jigen Peng and Shigang Yue, \IEEEmembership{Senior Member,~IEEE}
	
\thanks{This work was supported in part by EU HORIZON 2020 Project STEP2DYNA under Grant 691154, in part by EU HORIZON 2020 Project ULTRACEPT under Grant 778062, in part by China Postdoctoral Science Foundation Grant 2019M662837 and in part by National Natural Science Foundation of China under Grant 11771347. \emph{(Corresponding author: Shigang Yue, Jigen Peng and Cheng Hu.)}}
\thanks{J. Zhao, H. Wang, N. Bellotto and S. Yue are with the Computational Intelligence Lab, School of Computer Science, University of Lincoln, Lincoln LN6 7TS, U.K. (email: syue@lincoln.ac.uk). H Cheng is with Guangzhou University and University of Lincoln (email: chu@lincoln.ac.uk). J. Peng is with the School of Mathematics and Information Science,
	Guangzhou University, Guangzhou 510006, China (email: jgpeng@
	gzhu.edu.cn).}}

\markboth{IEEE Transactions on NEURAL NETWORKS AND LEARNING SYSTEMS}%
{Shell \MakeLowercase{\textit{et al.}}: Bare Demo of IEEEtran.cls for IEEE Journals}

\maketitle

\begin{abstract}
Collision detection is one of the most challenging tasks for Unmanned Aerial Vehicles (UAVs). This is especially true for small or micro UAVs, due to their limited computational power.
In nature, flying insects with compact and simple visual systems demonstrate their remarkable ability to navigate and avoid collision in complex environments. A good example of this is provided by locusts. They can avoid collisions in a dense swarm through the activity of a motion based visual neuron called the \textit{Lobula Giant Movement Detector} (LGMD). 
The defining feature of the LGMD neuron is its preference for looming. As a flying insect's visual neuron, LGMD is considered to be an ideal basis for building UAV's collision detecting system. However, existing LGMD models cannot distinguish looming clearly from other visual cues such as complex background movements caused by UAV agile flights. 
To address this issue, we proposed a new model implementing distributed spatial-temporal synaptic interactions, which is inspired by recent findings in locusts' synaptic morphology. We first introduced the locally distributed excitation to enhance the excitation caused by visual motion with preferred velocities. Then radially extending temporal latency for inhibition is incorporated
to compete with the distributed excitation and selectively suppress the non-preferred visual motions.
These spatial-temporal competition between excitation and inhibition in our model is therefore tuned to preferred image angular velocity representing looming rather than background movements with these distributed synaptic interactions. Systematic experiments have been conducted to verify the performance of the proposed model for UAV agile flights. The results have demonstrated that this new model enhances the looming selectivity in complex flying scenes considerably, and has potential to be implemented on embedded collision detection systems for small or micro UAVs.
\end{abstract}

%[A general LGMD work flow is consisted of 3 stage of image process, (1) motion information extraction and image preprocess, (2) simulated synaptic interaction, which is the defining stage of LGMD model that can extract information of looming objects. (3) output signal process (to reduce noise and enhance the needed information). Most of the modelling works intended to improve the performance from stage (1) and (3), due to the lack of knowledge of the synaptic structure.]

\section{Introduction}
%Several terms needed to be explained:

%Looming stimuli: objects approaching on a collision course or their simulation on a screen.

%retinotopy: 

%Neural Morphology: 

%presynaptic
%postsynaptic

\IEEEPARstart{A}{utonomous} flying robots or unmanned aerial vehicles (UAVs), especially small and micro aerial vehicles (MAVs), have increasingly displayed considerable potential for serving human society as a result of their flexibility of flight.
However, autonomous micro aerial vehicles (MAVs) remain unable to fly automatically and perform tasks safely. One of the reasons is that they have not been equipped with efficient collision detection capabilities. Traditional technologies of collision detection, such as laser~\cite{bachrach2009autonomous}, ultrasonic~\cite{yu2015sense}, and Simultaneous Localization and Mapping(SLAM)~\cite{Steder2008Visual} are computationally expensive, or greatly rely on objects texture and physical characters such as its ability to absorb and/or reflect light. These impediments make these methods unsuitable for MAVs. On the other hand, vision sensors can capture rich information of the real world and consume less power. 
However, exploiting the abundant information comes with a cost which, in this case is a demand for an efficient algorithm to extract task-specific features for collision detection. 
%[In collision detection, visual motion based algorithm is more computationally efficient than feature recognition in terms of acquiring dynamic information.] %引用 EMD， LGMD, DVS, feature recognition
%However, \textbf{Agile flight} of a UAV brings ego-motion of the camera which leads to confusing false positive in visual motion based algorithms %(This difficulty will be further explained later).
%Hence, meeting the needs of autonomous MAVs to discriminate critical collision information with limited computing resources is highly significant and challenging.
%\begin{figure}[t]
%	\centering
%	\includegraphics[width=0.95\linewidth,height=0.2\textheight]{Figures/Pitching_Example}
%	\caption{Attitude motions of a UAV (\textit{e.g.} a quadcopter) during \textbf{agile flight}. Being different from ground robots, the agile flight of a UAV is accompanied by attitude motions which lead to ego-motion of the camera and cause severe background noise to the captured images.}
%	\label{fig:pitchingexample}
%\end{figure}

Nature has demonstrated many successful solutions for dynamic collision detection. For example, locusts can fly with agility in a swarm of millions whilst avoiding collision. Their remarkable collision avoidance relies on a visual motion sensitive neuron: the Lobula Giant Movement Detector (LGMD)~\cite{rind1996neural}. The LGMD neuron has a characteristic preference for looming obstacles (i.e. objects approaching on a directly collision course) other than translating or receding objects, which makes it an ideal model for detecting collisions automatically~\cite{Rind2002Motion, fu2019review}. 
Although some of the LGMD inspired models have been successfully embodied on mobile robots for collision detection~\cite{hu2014development,hu2016bio}, and we also showed that a LGMD can detect collision in simple flying scenes with a constant flying speed~\cite{zhao2019lgmd}. However, these LGMD models cannot work when it comes to agile flights in complex environments.

%Considering locusts can avoid collision with amazing fly-ability, the failure to mimic its character indicate that existing models have not successfully fulfilled the preference for looming.

%cannot achive effective preference for looming in agile flights

\begin{figure}[t]
	\flushleft
	\subfloat[Grey samples]
	{
		\includegraphics[width=0.9\linewidth, height=0.05\textheight]{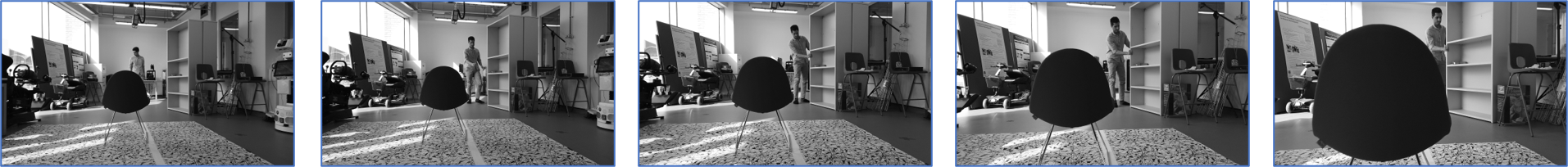}
	}
	\flushleft
	\subfloat[P layer samples (image motion)]
	{
		\includegraphics[width=0.9\linewidth, height=0.05\textheight]{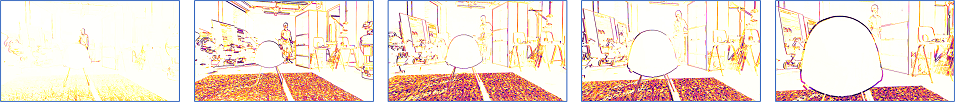}
	}
	\flushleft
	\subfloat[Normalised output membrane potential]
	{
		\includegraphics[width=0.95\linewidth,
		height=0.15\textheight]{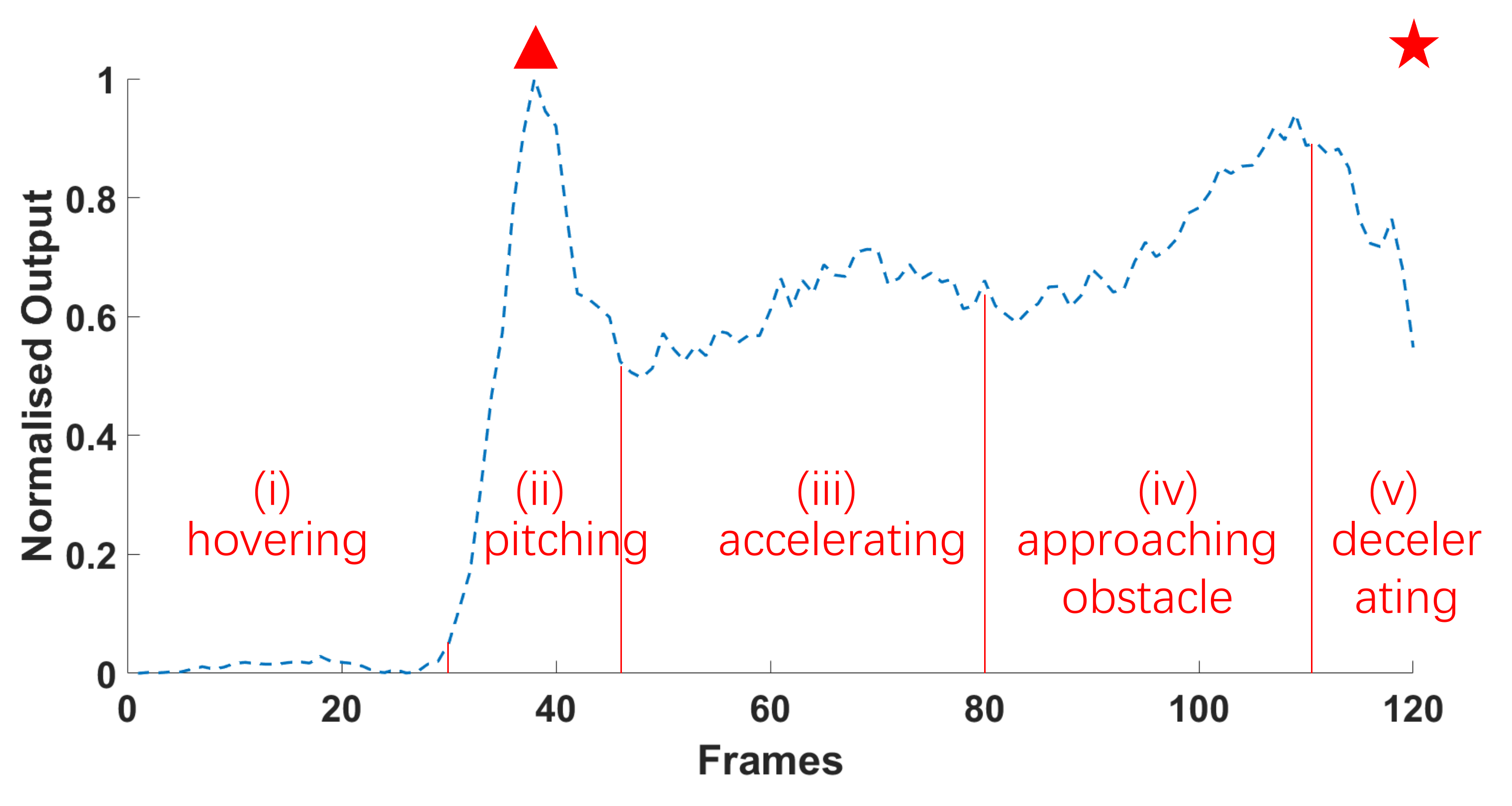}
	}
	\flushleft
	\subfloat[Typical response of locust's LGMD to looming stimuli]
	{
		\includegraphics[width=0.95\linewidth, height=0.12\textheight]{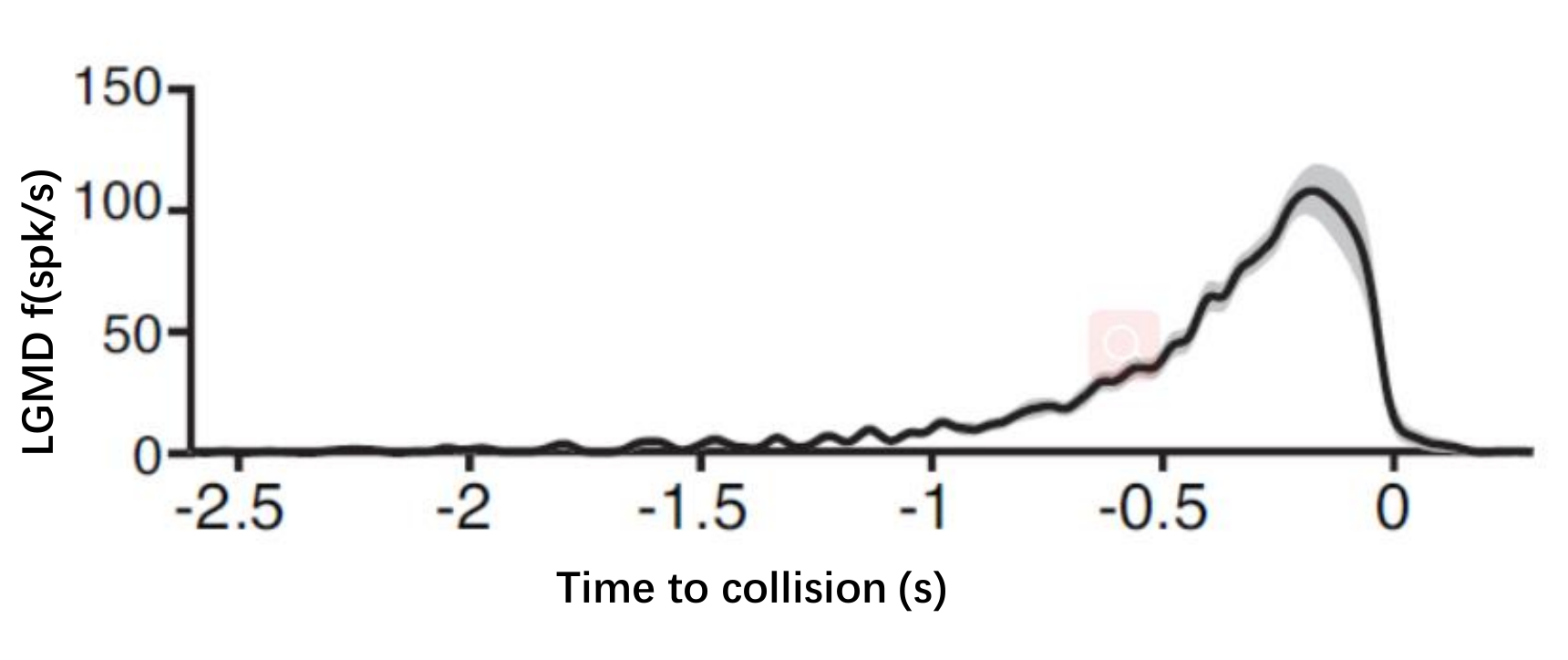}
	}

	\caption{An example of the conundrum during agile flight to a previous LGMD model~\cite{zhao2018bio}. The input video is collected while the UAV is flying towards the chair in a collision case under program control. \textbf{(a)} Input grey samples of the first-person-view (FPV) video. \textbf{(b)} Image motion samples as extracted from P layer. \textbf{(c)} Normalised membrane potential (MP) of the model. Red triangle: the first peak of false positive, red star: the collision point. The flight experienced 5 periods: (i) hovering, (ii) pitching, (iii) accelerating, (iv) approaching obstacle (looming) (v) Program controlled decelerating (to avoid hardware damage). It is seen that background noise leads to false positives in pitching (ii) and accelerating (iii) period. \textbf{(d)} A comparison of the locust LGMD neuron's response when facing looming stimuli. Y axis is LGMD's average spikes/second which is related to membrane potential (MP) and the figure is adapted from~\cite{dewell2018biophysics}.
	}\label{Fig:Conundrum}
\end{figure}

Essentially, \textbf{agile flight} is inevitable for flying robots in complex environments, as quick acceleration and ego motions are critical to maintain adequate efficiency. However, it often causes severe background noise to challenge the onboard vision sensor. A conundrum thus exists whereby the quadcopter needs to fly quickly to sustain efficiency, but this results in attitude motion and leads to false positive alarm from motion sensitive models. As an example of this difficulty, Fig.~\ref{Fig:Conundrum} present the performance of our previous LGMD model~\cite{zhao2018bio} during the flight towards a chair in a collision case. The LGMD model triggers non-negligible false positive alarms during the pitching and accelerating periods. 
Note these attitude motions in periods (ii) and (iii) of Fig.~\ref{Fig:Conundrum}(c) are crucial for achieving efficient agile flights by UAV's, and the fitting of pan-tilts is not feasible for power limited MAVs. To achieve satisfactory performance in efficient flight, there is a strong demand for improving LGMD model to cope with the characteristics of agile flights and maintain its looming selectivity in these scenes. 

%A typical LGMD model involves a few layers, including (1) motion information extraction and image preprocessing (corresponding to photoreceptors), (2) simulated synaptic computation (corresponding to synaptic excitation, inhibition and their interactions), which is the key stage for discriminating looming, and (3) output signal post-processing or enhancement (corresponding to grouping cells, FFI and LGMD output). 
A typical LGMD model consists of a few layers, including Photoreceptors, excitation, inhibition, synaptic summation and Feed forward inhibition and a LGMD neuron ~\cite{rind1996neural, yue2006collision}. The synaptic interactions in these models are simple, for example, the excitation is one-to-one connected from photoreceptor layer to summation layer, the lateral inhibition is  spatially distributed but not considering temporal distribution.

 Recent researches on locusts' LGMD neuron indicated that both excitatory~\cite{zhu2018pre} and inhibitory~\cite{rind2016two} synaptic pathways are exquisitely structured and locally distributed to interact with neighbours. The retinotopic mapping seems to play an important role in LGMD's preference for looming~\cite{dewell2019active}, which has been underestimated in previous models. These findings support the assumption that LGMD's synapses do discriminate spatial-temporal patterns which are embedded across its thousands of synaptic inputs~\cite{dewell2018biophysics}. 
 
Inspired by these findings, we propose the distributed presynaptic connection (DPC) structure and incorporate novel morphological strategies, to implement a spatial-temporal filter on image angular velocity to cope with agile flights. The DPC structure involves novel strategies: (1) locally distributed excitation to enhance the excitation caused by visual motion with preferred velocities, and (2) radially distributed temporal latency for inhibition to compete with the distributed excitation and selectively suppress the non-preferred visual motions. We experimentally analysed how the spatial-temporal distributions contribute to LGMD's looming selectivity and demonstrated the proposed model performs well in agile flights.

In summary, the contributions of this paper are threefold:
\begin{enumerate}
	\item We proposed a new LGMD model with locally distributed excitation for enhancing image motion with preferred angular velocity (on image) to better cope with agile flights.
	\item Temporal distribution is considered and defined in a radially extending manner to compete with the distributed excitation and form the spatial-temporal filter for looming cues.
	\item We demonstrated in experiments that the proposed model exhibited distinct preference for looming objects in agile flights and therefore is competent for collision detection for UAVs. %which considerably enhances the looming selectivity against other interfering stimuli.
\end{enumerate}
The remainder of this paper is organized as follows. In section II, we review related work on the LGMD models and UAV collision detection. In Section III, our proposed model is formally described with formulations. In Section IV, materials and the experimental setup are described.
In Section V, the results of experiments are presented and discussed. Section VI concludes the paper.

\section{Related Work}
In this section, we review the traditional methods used in UAV collision detection. This is followed by a consideration of bio-inspired approaches. Finally, recent relevant biologic researches on LGMD collision detector is introduced.
\subsection{Traditional UAV's Collision Detection}
For UAVs, visual collision detection systems can be categorised into two strategic approaches. One is to sense depth and escape when an obstacle is located at a given distance. The other one is to utilize monocular features in the recognition of obstacles or potential risks of collision, without sensing depth. 

The first approach requires %the controller to receive precise depth 
real-time knowledge about the 3D environment. This can be obtained with the use of a stereo-camera~\cite{hrabar2005combined}, LIDAR~\cite{richter2018bayesian,richter2016polynomial}, optic flow based distance maintenance~\cite{sabo2016bio,keshavan2015autonomous} or SLAM~\cite{Steder2008Visual}.
%and even monocular camera~\cite{saxena2006learning} is possible to get depth information
Such methods are commonly used in UAVs that have sufficient power. However, they cannot discriminate between objects. They simply compute the distance to every object in the whole field of view (FoV), which requires excessive calculation power.
Interestingly, in nature only higher species or predators present depth-based detectors, whereas insects tend not to be able to perceive "depth", due to limitations resulting from the spacing between their eyes and the lack of overlap in their binocular field. As a consequence, they sacrifice accuracy for a gain in efficiency by making use of basic monocular visual cues which allow them to sense the risk of collision or danger.
%Optic flow based distance maintenance is not available for strait-on collisions.

%How to detect collision, is a question for all creatures. One complete solution is to detecting the distance to objects. If you keep distance to objects, you will not get into collision. But it is not easy to get depth information in all the cases. Bat use echo, and human has stereo eyes. But the insects are observed to remain collision sensitive ability even with monocular vision cue. What's the mystery of visual based collision detection? What are the features that can be used to identify a collision? Recent studies on locusts vision has give us a glimpse under the vest.
The second strategy is to make plenty use of monocular visual features, including but not limited to colour, size, and image motion. For example, one can use a Speeded Up Robust Features (SURF) algorithm to recognise objects by feature points, and avoid the objects in the frontal area~\cite{aguilar2017obstacle,carloni2013robot}.
However, recognising the object is not necessary for nor sufficient for detecting potential collisions. It is indirect and therefore neither robust nor efficient.
Based on the observation that images of looming objects expand rapidly and in a non-linear way in the run-up-to collision, some researches attempt to detect the expansion of image edges in order to identify approaching obstacles. 
For example, a SURF algorithm and template matching method have been employed to detect the relative-expansion of a looming object\cite{mori2013first}. 
In addition, a Scale-invariant feature transform (SIFT) algorithm and template matching have been used to detect the relative-expansion \cite{al2017obstacle}.
Generally, these traditional recognition methods demand considerable computational power for handling data in complex dynamic scenes.

\subsection{Bio-inspired Collision Detection}
Animals have evolved with efficient sensor systems specialised for their living environments, and many insects are equipped with a designated visual system for flight.
For example, optic flow is such an insect (fly)-inspired method for visual motion perception. The field of optic flow can be used for estimating UAV's ego-motion~\cite{ZuffereyEgo} or collision detection by maintaining the balance of bilateral optic flow~\cite{sabo2016bio,serres2017optic}. However, it is not effective for detecting head-on collisions. In a complementary approach, optic flow has been combined with the ability to detect expansion. This led to a head-on collision detection system based on divergent optic flow~\cite{ZuffereyFly,stevens2018vision}.

LGMD is a visual neuron found in the locust's vision system to provide collision detection. Its characteristic looming selectivity largely arises within its dendritic fan~\cite{Graham2002How}. Because of its compact size and specialised sensitivity to looming objects, computational researches has modelled and applied LGMD network to robots for head-on collision detection~\cite{yue2006collision,hu2014development}. In engineering application, a typical LGMD network can be treated as 3 stages of image process, i.e. (1) motion information extraction and image preprocess, (2) simulated synaptic computation, including local interaction and global summation. This is the key stage of LGMD model that underlies the ability to discriminate looming information. (3) output signal process (to reduce noise and enhance the interested information). A successful hypothesis for modelling the defining feature of LGMD model (stage 2 process) is proposed by Rind~\cite{rind1996neural} in 1996, which point that the LGMD neuron can extract fast moving edges through the "critical race" \footnote{i.e. Excitation caused by an edge motion must move fast to escape from the impact of laterally distributed inhibition, otherwise, it will lose the race and cannot reach the threshold to activate downstream synapses} formed by excitation and lateral inhibition. Based on this hypothesis, it can be deduced that a looming object can be identified by its fast moving edges and the angular size it occupied on the retina. 
After that, many researches developed this model through image preprocessing or post-processing. For example, Yue and colleagues\cite{yue2006collision} introduced extra grouping layer to enhance the clustered output and improve the performance. Q Fu \cite{fu2019TCYB} proposed on and off pathways in the 1st stage and focused on dark objects in light background; Another paper~\cite{fu2020Improved} feed back global intensity to mediate the inhibition weights in order to acquire adaptive sensibility according to different background complexity; Meng~\cite{meng2010modified} introduced additional cells in the 3rd stage to acquire change rate of the converged output so that the model can predict approaching or receding movements. He~\cite{HeLei2020moment} introduced image moment in preprocessing to enhance the resistance against ambient light change. But these researches mainly contributed to stage 1 or 3, because it is still not exactly known how the synapses corporate to achieve looming information extraction and it is hard to propose new biological plausible model to mimic the synaptic process. Besides, there is another hypothesis about the synapitc interactions claimed that the synapitc interaction can involve multiplication~\cite{gabbiani1999many}, and based on this, Badia~\cite{Badia2007Blimp} modeled the presynaptic layer with (Reichart correlator based EMDs~\cite{reichardt1961autocorrelation}) to detect expanding edges. However, their model requires a unique preprocessing to predict expanded image, which has no evidence supported in biology. Despite this, because it is another related work to model the presynaptic interactions, we will also compare this model in Sec~\ref{subsec:FPVVideo} (Fig.~\ref{Fig:ComplexFlight}).

Unfortunately, in agile UAV flights, the complex dynamic image motions generate spurious signals and will challenge the existing LGMD models with false positives. These false positives can hardly be solved by preprocessing or post-processing, demonstrating that the synaptic interactions (in stage 2) of existing models are too simple to discriminate spatial-temporal pattern as we expected.

%Therefore, there is a demand to enhance the LGMD model with the expectation to maintain its preferential response to looming objects against fast interfering signals.

\subsection{Emerging Biological Findings about LGMD}
Thanks to technology development, recent biology can go further to explore the interneuron connections of the LGMD's dendrite. 
Recent biological researches have highlighted the importance of the retinotopic reciprocal connections within the dendritic area. It is also reported that both excitatory and inhibitory presynaptic connections have a degree of overlap~\cite{zhu2016fine,rind2016two}, which is different from existing LGMD inspired models~\cite{hu2016bio,fu2019TCYB,fu2019review,zhao2018bio}. Zhu suggested that the distributed excitation increases in response to coherently expanding edges~\cite{zhu2018pre}. To conclude, these findings indicate that dendrites receive finely distributed retinotopic projections from the photoreceptors and interact with neighbouring synapses before they converge. The locally distributed interaction is not as simple as previously assumed~\cite{rind1996neural}, but potentially forms a filter to discriminate spatial temporal patterns that are mapping across its dendritic fan.

In this paper, we update the model with such a presynaptic layer, which contains locally distributed excitatory and inhibitory reciprocal connections. 
In the proposed model, the spatial temporal structure of synaptic activity is determined by an overall spatial-temporal distribution in the distributed presynaptic connection (DPC) layer. 
In experiments, selective response to images with different angular velocity is initiated after the retinotopic mapping in the DPC layer, and before the postsynaptic inhibition (the FFI), demonstrating the DPC process successfully simulated the preference for looming of LGMD neuron. 
%In developing a novel LGMD-inspired collision detector, our work is different from that reported previously~\cite{bermudez2007fly}, where nonlinear preprocessing of the image and multiplicative reciprocal connection were involved. Our model extracts looming visual cues using linear reciprocal connections only. Compared to previous research~\cite{yue2006collision,zhao2018bio}, the looming selectivity in our model is considerably enhanced and is able to be tuned with flexibility using the spatial-temporal distribution in the DPC layer. 
As a result, our model exhibits greatly enhanced robustness in complex scenes. Compared to traditional visual methods, the proposed presynaptic filter is based on linear processing of luminance change, which makes it computationally efficient and endows with the potential to be applied on micro embedded systems.

\section{Model Description}
This section formulates the proposed distributed presynaptic connection based LGMD model (named \textit{`D-LGMD'}). 
Considering the neural process is continuous, the whole model is re-formed in continuous integral format, but the contribution of this paper is focused on stage (2) process of LGMD.
Besides, in order to retain the looming selectivity during complex background motion, some modification has been made to the threshold process in~\ref{subsec:FFI-GD}.
\subsection{Mechanism and schematic}
\begin{figure}
	\centering
	\includegraphics[width=0.95\linewidth,
	height=0.15\textheight]{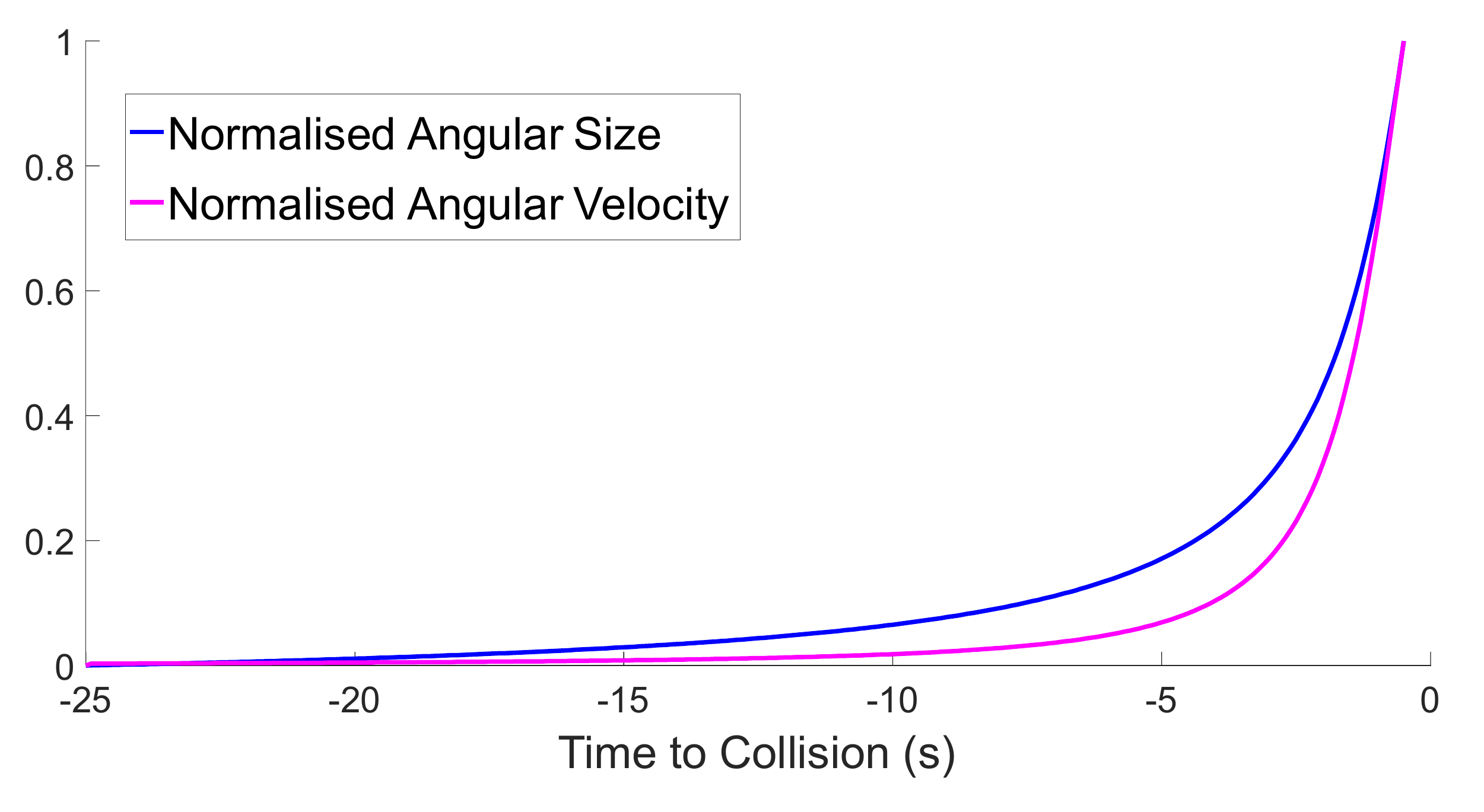}
	\caption{Ideal object angular size and angular velocity on image as a function of time during a looming incident. Both curves are nonlinear. The setting of a threshold on angle or angular velocity may be the strategy used by insects to identify collision cases~\cite{gabbiani1999computation}. The proposed methods develop the use of nonlinear angular velocity for collision detection.}\label{Fig:Ideal_Looming_Change}
\end{figure}

As determined from geometric analysis, the image of an ideal looming object shows a sharp nonlinear expansion as the object nears the collision point. The angular size and angular velocity of the image both increase non-linearly~\cite{Gabbiani1999Comput} as shown in Fig.~\ref{Fig:Ideal_Looming_Change} . 
The non-linear angular velocity symptom of a looming object is very unlikely to be produced by other sources of visual stimuli, such as receding or translating objects. Therefore, we aims to form a spatial-temporal filter in DPC layer to discriminate angular velocities of images on the retina. Following the idea of “critical race” by Rind~\cite{rind1996neural}, The DPC layer boosts signals derived from fast expanding edges of looming objects and eliminates interfering stimuli caused by other visual sources. Different from previous models~\cite{yue2006collision,zhao2018bio,fu2019TCYB}, the proposed DPC layer can achieve accurate image angular velocity preference through a combination of locally distributed excitation and the spatial-temporal race against inhibition. %Excitation of an object must win the race to reach the threshold in later layers. The faster the edge moves over the camera, the more excitatory units stand out. Once excitation of the coherently expanding edge exceeds the offsetting impact of inhibition, excitatory units mutually enhance because of the distributed reciprocal connection. The discriminative response to different angular velocity is tuned by the distribution functions which are pre-defined and remain constant during the computing process.

\begin{figure}
	\centering
	\subfloat[]
	{
		\includegraphics[width=0.95\linewidth,
		height=0.25\textheight]{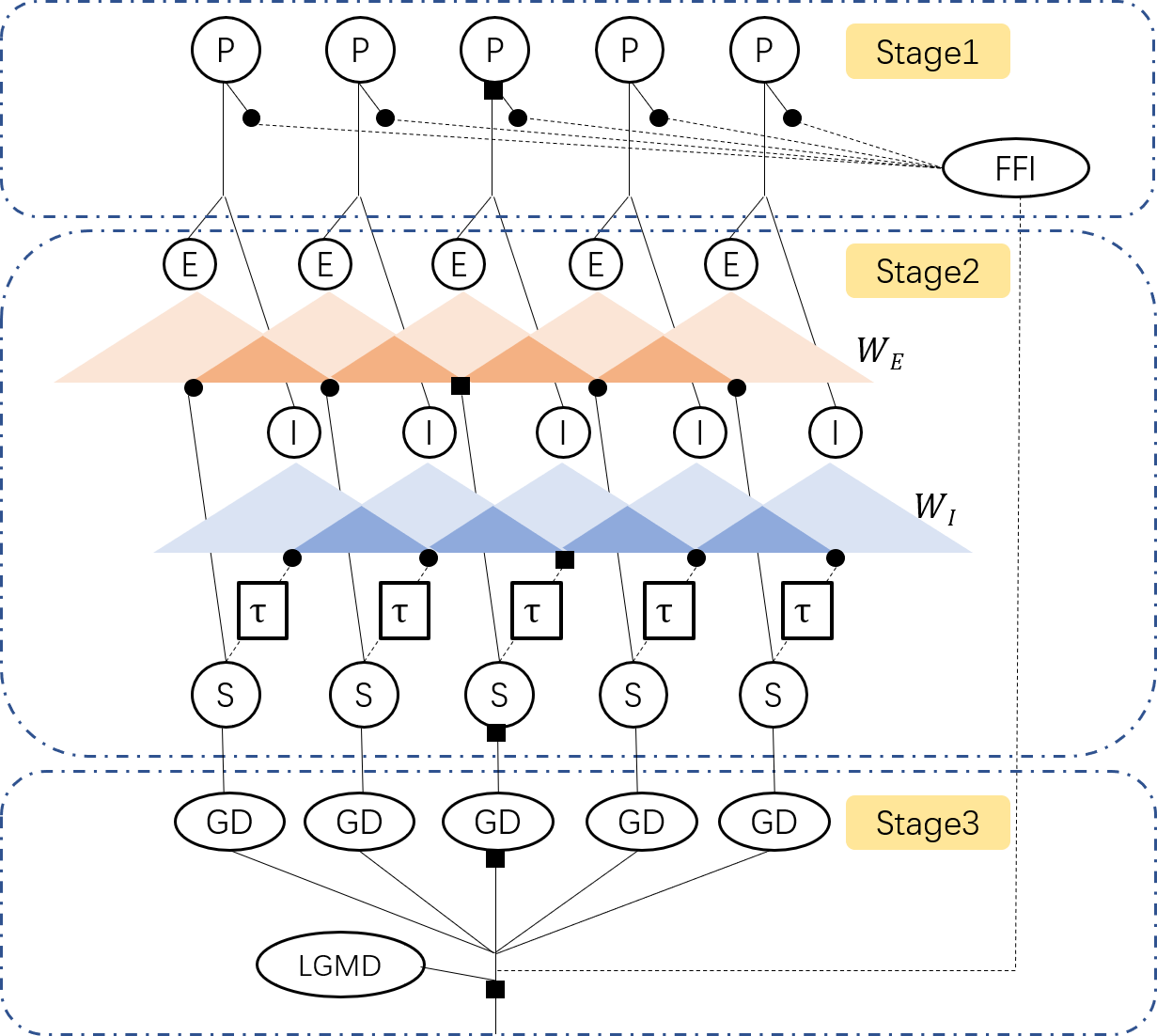}
	}
\\
	\centering
	\subfloat[]
	{
		\includegraphics[width=0.95\linewidth,
		height=0.18\textheight]{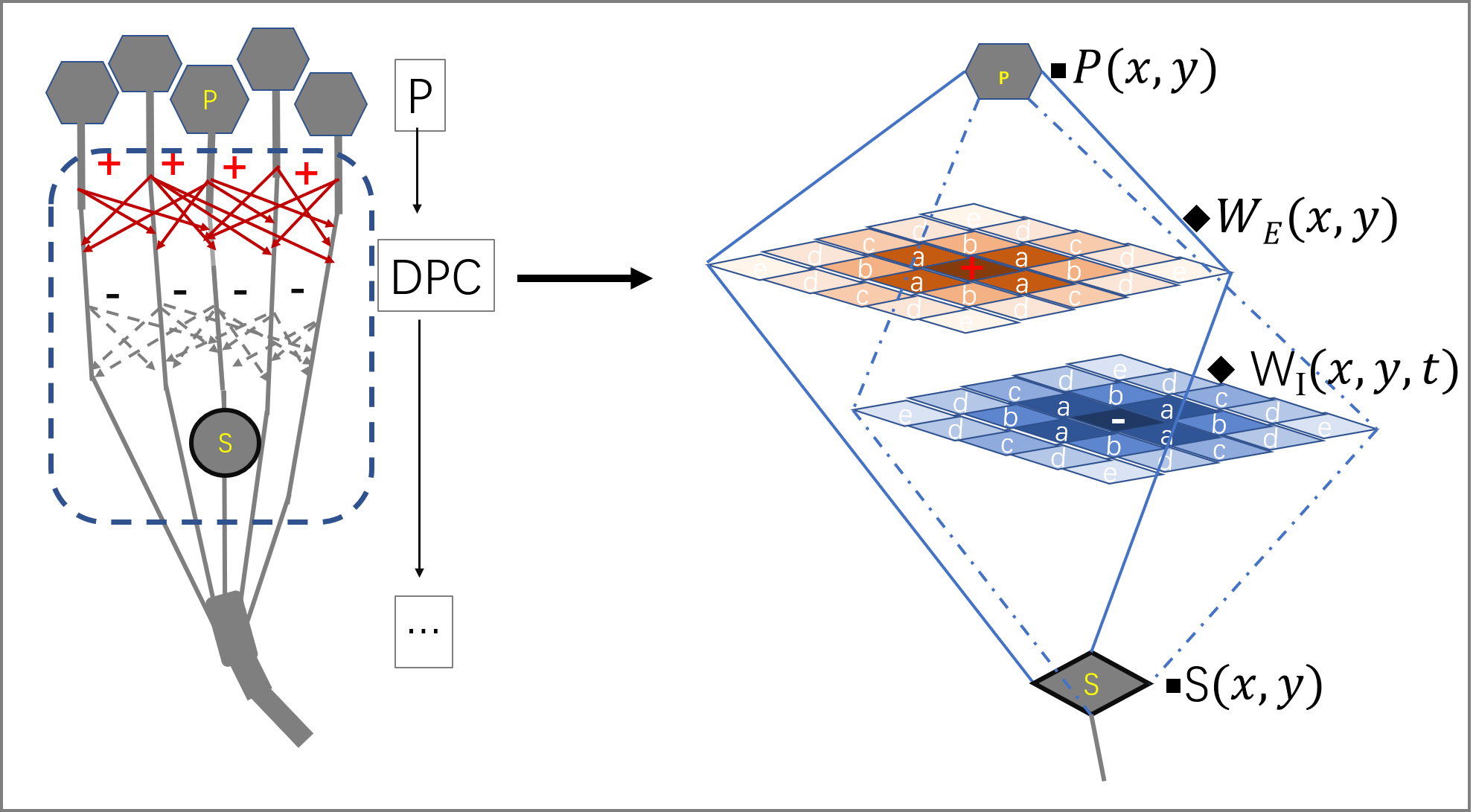}
	}
	
	\caption{Model schematic. \textbf{(a)} schematic of the proposed D-LGMD model. P: photoreceptors, E: excitation, I: inhibition, S: sum, G: group and decay, FFI: Feed forward inhibition. Our contribution focus on the second stage especially the spatial-temporal morphology of the locally distributed mapping, which is indicated by the orange ($W_E$) and blue shadow ($W_I$). \textbf{(b)} illustration of DPC process (from P to S) for a single pixel (x,y). "+" denotes excitation, "-" denotes inhibition. Both excitation and inhibition are mutually connected with neighbouring synapses, and the connectivity is ruled by distribution functions $W_E(x,y)$ and $W_I(x,y,t)$. Note the inhibition latency $\tau$ is also defined by $W_I(x,y,t)$.
	}\label{Fig:Ideal_LGMD}
\end{figure}

%\begin{figure}
%	\centering
%	\includegraphics[width=0.95\linewidth,
%	height=0.15\textheight]{Figures/Ideal_Looming_Angular_1}
%	\caption{Ideal object angular size and angular velocity as a function of time during a looming incident. Both curves are nonlinear. The setting of a threshold on angle or angular velocity may be the strategy used by insects to identify collision cases~\cite{gabbiani1999computation}. This paper demonstrates the use of nonlinear angular velocity for collision detection.}\label{Fig:Ideal_Looming_Change}
%\end{figure}
A schematic of the proposed D-LGMD model is presented in Fig.~\ref{Fig:Ideal_LGMD}. The D-LGMD model is comprised of 3 stages of image processing: 1 motion information extraction (photoreceptors), 2 synaptic local interaction and global summation, 3 output feed back or forward and feature enhancement. 
The photoreceptors extract image motion, and divides into excitatory and inhibitory pathways. Then the synapses interact with neighbours through the morphologic mapping in DPC layer. after that, inhibition and excitation sum up and will be threshold after grouping and decay (GD).
%the photo receptors, the distributed presynaptic reciprocal connection (DPC), and the grouping and decay area (GD).
Additionally, the feed forward inhibition (FFI) component, as a side pathway of postsynaptic inhibition~\cite{santer2004retinally} mediates the threshold to regulate output MP within a dynamic range. This function is termed FFI mediated grouping and decay (FFI-GD). 

\begin{figure}
	\centering
	\includegraphics[width=0.95\linewidth,
	height=0.18\textheight]{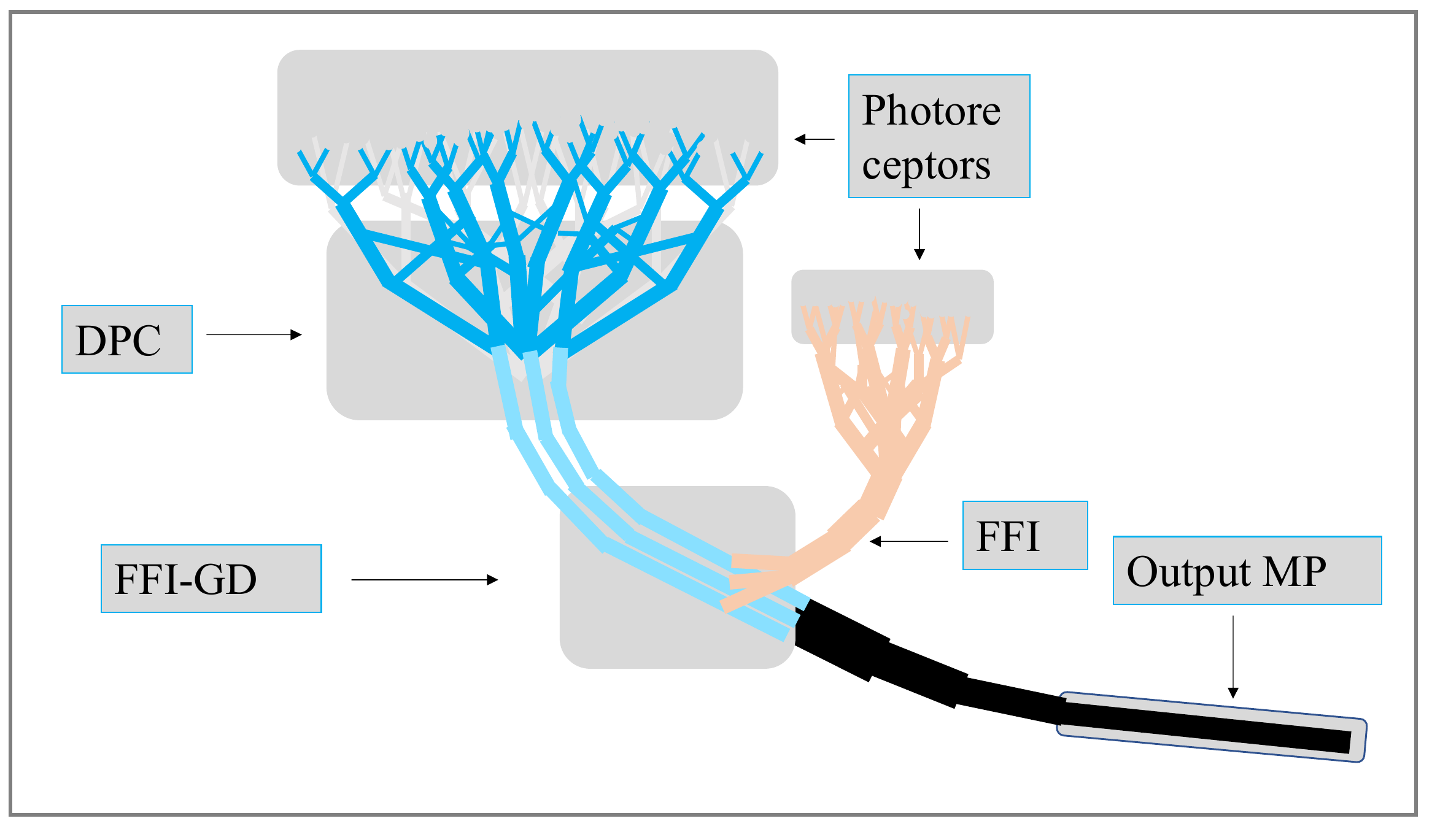}
	\caption{D-LGMD neural model. DPC: distributed presynaptic connection, FFI: feed forward inhibition, FFI-GD: FFI mediated grouping and decay, MP: membrane potential}
	\label{fig:model}
\end{figure}

Finally, a single output terminal, of which the membrane potential (MP) reflects the threat level of collision in the whole FoV, instructs downstream motion systems to avoid collisions. 
The morphology of the proposed D-LGMD is shown in the neural model in Fig.~\ref{fig:model}.
In summary, there are 3 main differences compared to previous LGMD models:

\begin{enumerate}
	\item Excitatory and inhibitory synaptic pathways are both locally distributed and interact with neighbours. They compete in space but also boost coherently expanding edges if they win the competition.
	\item The inhibitory latency is radially distributed and increases as the transmission distance extends. This pattern of latency boosts the competition between excitatory and inhibitory synaptic afferents.
	\item In the side pathway, FFI no longer switches off the output MP. It mediates the decay threshold after grouping. This new mechanism keeps the output MP in a dynamic range and enables the detector to remain sensitive to looming stimuli in a rapidly changing FoV (as occurs during attitude motion). 
\end{enumerate}

%\begin{enumerate}
%	\item \textbf{Excitatory and inhibitory synaptic mappings are determined by spatial-temporal distributions in the DPC layer which pre-defines the preferred angular speed clearly. Based on this, D-LGMD can easily ignore insignificant cues below the threshold and produce a nonlinear response to looming objects. The selective response with non-linearity not only increases the robustness against background noise but also boosts the quick response to collision when looming objects become prominent.}
%	\item In the side pathway, FFI no longer switches off the output MP. It mediates the decay threshold after grouping. This new mechanism keeps the output MP in a dynamic range and enables the detector to remain sensitive to looming stimuli in a rapidly changing FoV (as occurs during attitude motion). 
%\end{enumerate}

%\begin{figure}
%	\centering
%	\includegraphics[width=0.95\linewidth,
%	height=0.18\textheight]{Figures/DPC_Schematic_1}
%	\caption{Illustration of the DPC process for a single pixel. "+" denotes excitation, "-" denotes inhibition. Both excitation and inhibition are mutually connected with neighbouring synapses, and the connectivity is ruled by distribution functions $W_E(x,y)$ and $W_I(x,y,t)$. [There is a brand new mechanism from the model in ~\cite{zhao2018bio}: The activation of each facet no longer provides point-to-point like stimuli, but will now stimulates a region of synapse.][Another difference is inhibition also passes to its direct counterpart in the next layer, which in represent for self-inhibition mechanism.]}
%	\label{fig:DPC_Schematic}
%\end{figure}

\subsection{Photo-receptor layer}
The first layer of the proposed model is a photo receptor layer (P layer). To behave as a motion sensitive visual model, the input layer monitors changes in the absolute luminance hitting each pixel:
\begin{equation}\label{qt:1}
P(x,y,t)= |L(x,y,t)-\int L(x,y,s)\delta(t-s-1)ds|
\end{equation}
where $\delta$ is the unit impulse function, $P(x,y,t)$ denotes the change in luminance of pixel $(x,y)$ at time $t$, and $L(x,y,t)$ refers to the luminance at time $t$. The P layer responds to all image motion equally and does not discriminate between backgrounds or foregrounds, translational, receding or looming movements.

\subsection{DPC layer}
\label{subsec:DPC}
\begin{figure}
	\centering
	\subfloat[]
	{
		\includegraphics[width=0.55\linewidth, height=0.15\textheight]{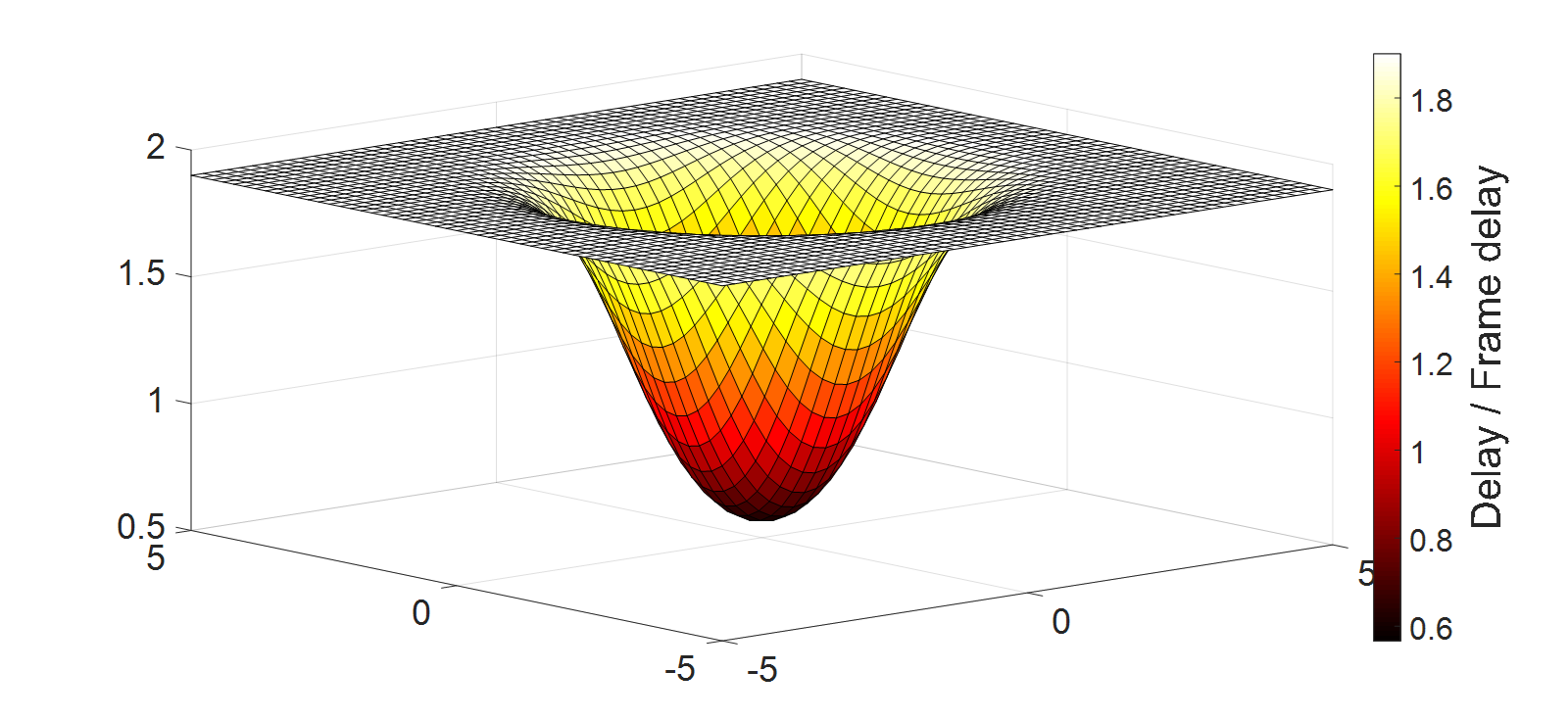}
		
	}
	\centering
	\subfloat[]
	{
		\includegraphics[width=0.4\linewidth, height=0.135\textheight]{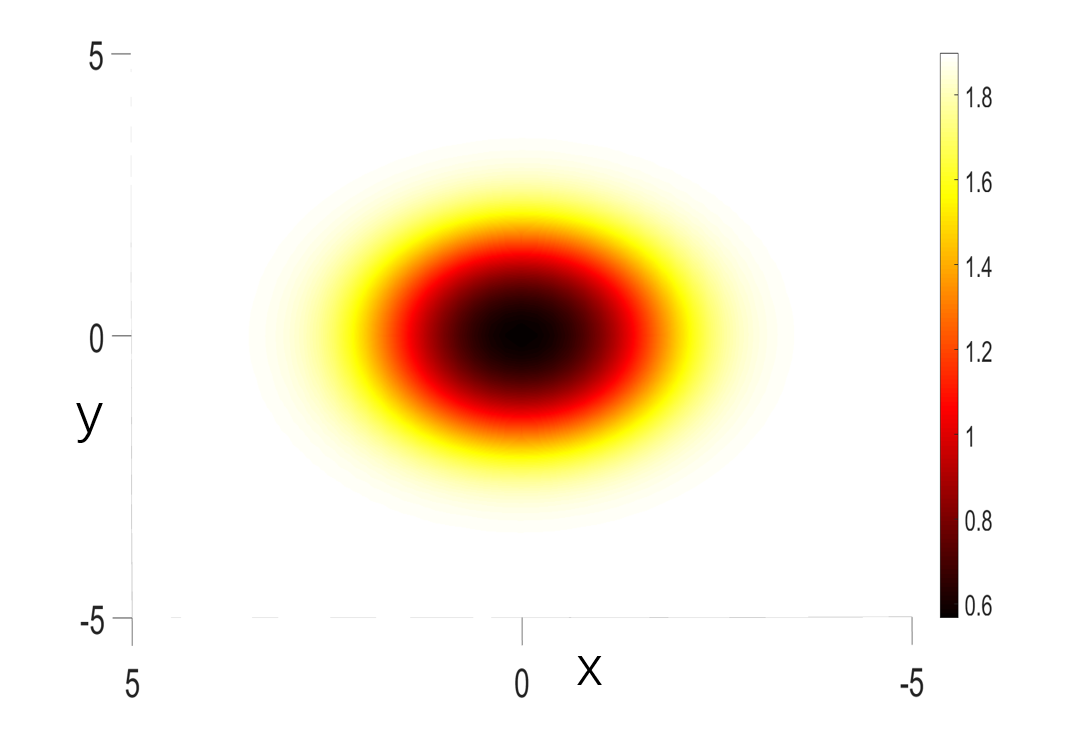}
	}
	\caption{An example of the radially extending temporal latency distribution $\tau(x,y)$. \textbf{(a)} kernel mesh of $\tau(x,y)$, \textbf{(b)} top view of the kernel. $\alpha,\beta,\lambda$ are set as $\alpha=-0.1,\beta=0.5, \lambda=0.7$.}\label{fig:delayht-kernel}
\end{figure}

\begin{figure}
	\flushleft
	\subfloat[Constant $h(t)=1$]
	{
		\includegraphics[width=0.9\linewidth, height=0.11\textheight]{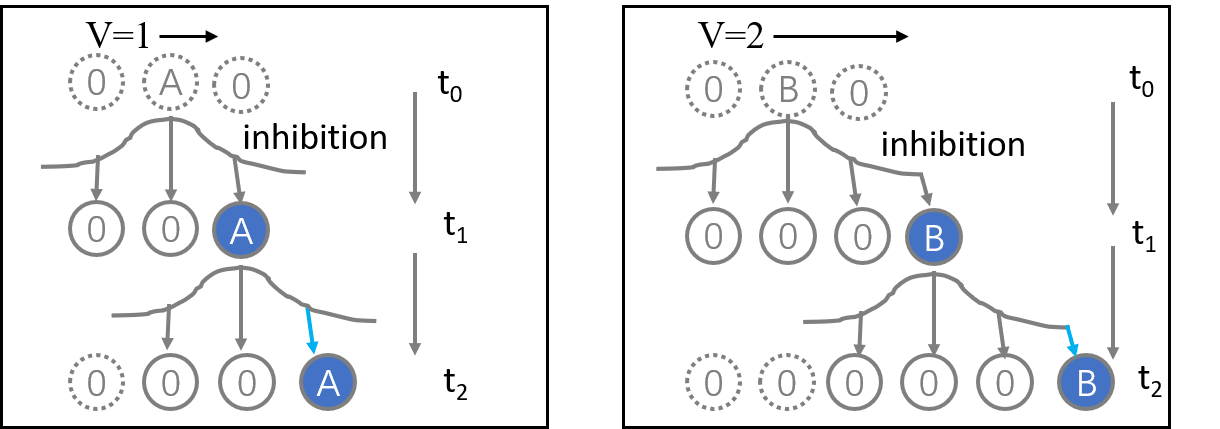}c
		\label{fig:Neural_constant_ht}
	}
	\flushleft
	\subfloat[Distributed $h(t)$]
	{
		\includegraphics[width=0.9\linewidth, height=0.11\textheight]{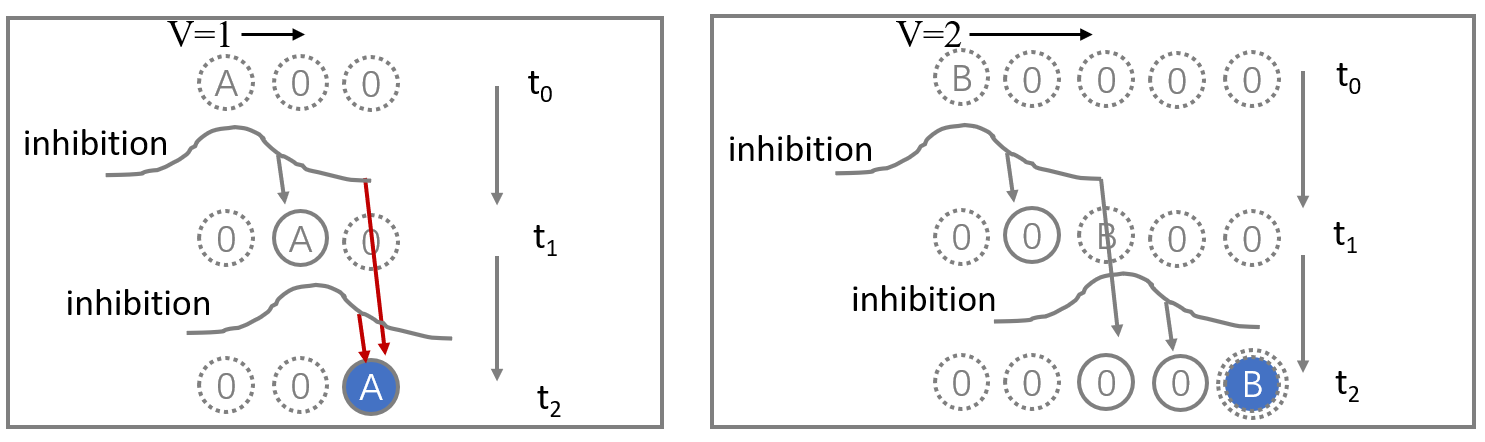}
		\label{fig:Neural_distributed_ht}
	}
	\caption{A contrast of inhibitory impact received by slower stimulus (A) and faster stimulus (B) under different time latency types. The time latency $\tau(x,y)$ is constant in \textbf{(a)} while radially distributed in \textbf{(b)} (with this in mind: a simple $\tau(x,y)$, 1 frame delay for 1 pixel distance and 2 frame delay for 2 pixel distance, is given as an example of radially distributed latency). In \textbf{(a)}, with constant latency, inhibition passes to the latest frame $t_2$ are "isolated" (indicated by blue arrows), no mater the stimulus moves slow (A) or fast (B). In \textbf{(b)}, with radially extending latency, inhibition passes to $t_2$ is accumulated at position A, where stimulus A receives "replicate" inhibitory impact (indicated by red arrows). Contrarily, stimulus B completely escapes from the range of the impact and therefore, it is released from inhibition. This demonstrates that a radially distributed latency $\tau(x,y)$ will selectively enhance the inhibition to slowly changing stimuli and let through the preferred (fast) stimuli.
	}\label{Fig:Neural_shematic_T}
\end{figure}

Below the P layer, image changes of the whole FoV are extracted, and only the information on moving edges is input to the subsequent DPC layer. The proposed DPC layer defines the second stage of D-LGMD process, which is the key stage that forms the looming selectivity. This layer enhances stimuli from images of looming or high speed objects and inhibits those from objects involved in lateral translation or from the background. Fig.~\ref{Fig:Ideal_LGMD} (b) illustrates the DPC process in pixel manner. Note both the excitation and inhibition are locally distributed.
%The ability to discriminate between different sources of visual stimuli is 
%\textbf{formed by the reciprocally connected and interacting neural pathways which facilitate interneuron excitation, interneuron inhibition and self-inhibition (i.e. inhibition also passes to its direct counterpart in the S layer). }
In relation to the current body of research, concerning the LGMD neuron in locusts, the characteristic of the DPC layer are consistent with the following principles:
%是否应该放到related work 里面？
\begin{enumerate}
	\item Both excitatory and inhibitory pathways are locally interconnected~\cite{rind2016two}.
	\item The strength of connection tapers along the diameter from the root towards the dendritic tip.~\cite{cuntz2007optimization}~\cite{cuntz2010one}.%不一定靠谱
	\item The time race between excitation and inhibition is essential for the preference to image angular velocity.~\cite{rind1996neural}.
\end{enumerate}
We have therefore constructed the DPC layer with 2 distributions for excitatory and inhibitory pathways. Furthermore, the time race between excitation and inhibition is integrated in distribution functions as follows:
\begin{align}
& E(x,y,t)=\iint P(x,y,t)W_E(x-u,y-v)dudv  \label{qt:2} \\
& I(x,y,t)=\iiint P(x,y,t)W_I(x-u,y-v,t-s)dudvds  \label{qt:3}
\end{align}
where $E(x,y,t)$ and $I(x,y,t)$ are excitation and inhibition at each pixel, respectively. $W_E$, $W_I$ are the distribution functions of excitation and inhibition, respectively. In relation to principle 3), $W_I$ contains distributions not only in the spatial domain but also in the temporal domain.
In agreement with the principle 2), a Gaussian kernel is chosen to describe the two distributions in the spatial domain.
\begin{equation}\label{qt:4}
\begin{cases}
W_E(x,y)=G_{\sigma_E}(x,y) \\
W_I(x,y,t)=G_{\sigma_I}(x,y)\delta(t-\tau(x,y))
\end{cases}
\end{equation}
In eq. (\ref{qt:4}), $\sigma_E$,\ $\sigma_I$ are standard differences of excitation and inhibition distribution (Note in application, another parameter $r$ will be involved to limit the size of the kernels\footnote{In application, $r$ can be smaller as long as it is plenty to possess the characters of the spatial-temporal mappings adequately}). $\tau(x,y)$ is the temporal mapping function of inhibitory pathways; The latency is distance-determined, and increases as transmission distance extends.
\begin{equation}\label{qt:5}
\tau(x,y) =\alpha + \frac{1}{\beta+\exp(-\lambda^2(x^2+y^2))}
\end{equation}
In eq. (\ref{qt:5}), $\alpha,\beta,\lambda$ are time constants. An example of the temporal latency distribution is shown in Fig.~\ref{fig:delayht-kernel} (note when $\alpha=\beta=\lambda=0$, $\tau(x,y)=1.$). Time latency is necessary to form the spatial-temporal race between excitation and inhibition, which has been well explained previously~\cite{rind1996neural}. In this research, we further put forth that the radially extending temporal distribution sharpens the output curve because it produces a gradient in temporal domain for inhibition. It thus selectively enhances the barrier against visual cues that are comparatively slow.
Fig.~\ref{Fig:Neural_shematic_T} elucidates this mechanism by comparing the inhibitory impact under constant latency (Fig.~\ref{Fig:Neural_shematic_T} (a)) and radially distributed latency (Fig.~\ref{Fig:Neural_shematic_T} (b)). In Fig.~\ref{Fig:Neural_shematic_T} (a), both stimulus A (the slow one) and stimulus B (the rapid one) at $t_2$ only receive an "isolated inhibition" (indicated by blue arrows) from the previous time stage $t_1$. In contrast, with distributed latency in Fig.~\ref{Fig:Neural_shematic_T} (b), stimulus A receives "replicate inhibition" (indicated by red arrows) from both the previous stage $t_1$ and the before-previous stage $t_0$, while stimulus B completely escapes from the inhibitory range. Therefore, the radially extending latency distribution allows the rapid/preferred stimuli to stand out in the model (more discussion about the advantage of radially distributed latency is given in Fig. ~\ref{Fig:Compare_h(t)}).

Subsequently, the distributed interconnections, excitation and inhibition, are integrated by a linear summation (note that inhibition has the opposite sign against excitation):
\begin{equation}\label{qt:6}
S(x,y,t)=E(x,y,t) - a \cdot I(x,y,t)
\end{equation}
In eq. (\ref{qt:6}), $S(x,y,t)$ is the presynaptic sum corresponding to each pixel at time $t$, and $a$ is the inhibition strength coefficient. Since additionally, synapses stimuli are not suppressed to give negative values, a Rectified Linear Unit (ReLU) is introduced:
\begin{equation}\label{qt:7}
S(x,y,t)= ReLu(0,S(x,y,t))
\end{equation}
Where $ReLu(x) = max (0,x)$. 

Using the above formulations, the DPC layer forms a spatial-temporal filter, of which the input comprises the motion of the image, and the output consists only of coherently edges from dangerous looming objects whose image on the retina moves relatively fast. Under the DPC manipulation, the temporal information of the image, which results from the latency between E and I pathways, cooperates with the spatial distribution and determines the character of preferred angular velocity. Specifically, the coherent excitations will mutually enhance but must move faster to escape from the rejection band otherwise they will be inhibited.
Therefore, only rapidly changing profiles of truly dangerous looming objects stand out after information has passed through the "spatial-temporal filter", while the stimuli caused by slowly translating objects or backgrounds are dramatically attenuated and are further eliminated by the threshold in the subsequent layer. The spatial-temporal distributions $W_E$, $W_I$, and $\tau(x,y)$ regulate the competition between excitation and inhibition, therefore, they are critical to shape the selectivity for objects with different angular velocities.
Importantly, since the DPC layer discriminates on the basis of angular velocity, if an object is close enough, to the extent that it occupies a large area of the retina and laterally translates at an extreme angular speed, it is also labelled as a "dangerous target". As a consequence, the model triggers an alarm. This character is consistent with empirical experiments which show locust behaviour towards a sudden translational movement~\cite{rind1992orthopteran}.

\subsection{FFI mediated grouping and decay}
\label{subsec:FFI-GD}
The output of the DPC layer is sent to a grouping and decay (GD) layer and to further reduce the noise and smooth the output. The grouping mechanism allows clusters of excitation from the DPC layer to easily pass to its corresponding GD counterpart and provides a greater MP output. This mechanism is implemented by multiplying the summation in the DPC layer with a passing coefficient $Ce$ as in eq. (\ref{qt:8}):
\begin{equation}\label{qt:8}
G(x,y,t)=S(x,y,t)\cdot Ce(x,y,t)
\end{equation}
Where $G(x,y,t)$ is the excitation that corresponds to each dendritic cell in the G layer and $Ce$ is an integration from its neighbourhood, and is given by:
\begin{equation}\label{qt:9}
Ce(x,y,t) = \iint_{\Omega} S(x,y,t)\cdot{k} dxdy
\end{equation}
where $k$ is a constant and $\Omega$ is the neighbourhood area. $\Omega$ is set to be a $4\times{4}$ matrix in this paper. G layer is followed by a threshold to filter out decayed signals:
\begin{equation}\label{qt:10}
\widetilde{G}(x,y,t)=\begin{cases}
G(x,y,t), & \mbox{if } G(x,y,t)\geqslant T_{de}(t) \\
0, & \mbox{otherwise}.
\end{cases}
\end{equation}
After that, a decay threshold $T_{de} (t)$ is involved to reduce interneuron output of inactive afferent. $T_{de} (t)$ is mediated by the side pathway postsynaptic inhibition, the FFI, and is given by:
\begin{equation}\label{qt:11}
T_{de}(t) = \frac{FFI(t)}{n_{cell}\cdot m}\cdot T_0
\end{equation}
%should we normalise the FFI in range (0~1) ?
$T_0$ is the baseline threshold, $n_{cell}$ is the total number of pixels in a single frame, m is a constant.
$FFI(t)$ is calculated by the previous image changes in the whole FoV, based on \ref{qt:12}:
\begin{equation}\label{qt:12}
FFI(t) = \iint|P(x,y,t-1)|dxdyds
\end{equation}
The side pathway FFI no longer switches off the output MP but mediates a threshold level for all single synaptic afferent according to the luminance change in the FoV. 
This new FFI-GD mechanism further suppress the edges caused by background motion and keeps the output MP within a dynamic range. Moreover, it preserves the ability of D-LGMD to work in complex and dynamic scenes.
\subsection{LGMD cell}
Finally, the MP of the LGMD cell $K(t)$ is the summation within the G layer derived from the whole FoV:
\begin{equation}\label{qt:13}
K(t)=\iint|\widetilde{G}(x,y,t)|dxdy
\end{equation}
If the MP $K(t)$ exceeds the threshold, a spike-like response is produced:
\begin{equation}\label{qt:14}
S_f^{spike}=\begin{cases}
1, & \mbox{if } K(t) \geqslant T_{MP} \\
0, & \mbox{otherwise}.
\end{cases}
\end{equation}
In application, an impending collision can be confirmed if successive spikes last consecutively for no less than $n_{sp}$ frames:
\begin{equation}\label{qt:15}
C_f^{LGMD}=\begin{cases}
1, & \mbox{if } \sum\limits_{f \!- \! n_{sp}}\limits^{f} S_f^{spike} \geqslant{n_{sp}} \\
0, & \mbox{otherwise}.
\end{cases}
\end{equation}
\begin{table}
	\centering
	\caption{Constant Parameters}
	\label{TB:parameter1}
	\begin{tabular}{|c|c|c|}
		\hline
		Paramter & Description & Value \\
		\hline
		k & Amplifying constant in Eq. \textbf{(\ref{qt:9})} & 1\\
		\hline
		$T_0$ & Threshold baseline in Eq. \textbf{(\ref{qt:11})}& 0.5\\
		\hline
		$m$ & Constant cofficient in Eq.\textbf{(\ref{qt:11})}& 0.4\\
		\hline
		$T_{MP}$ & LGMD spiking threshold in Eq. \textbf{(\ref{qt:14})}& 0.4\\
		\hline
		$n_{sp}$ & minimum spikes to alarm in Eq. \textbf{(\ref{qt:15})} & 2\\
		\hline
%		$T_0$ & 0.5 & $T_{MP}$ & 0.4\\
%		\hline
		%		\bottomrule
	\end{tabular}
\end{table}
The LGMD detector will generate an "avoid" command to the quadcopter if the spikes lasts $n_{sp}$ frames.
Regular parameters are listed in Table \ref{TB:parameter1}. These parameters are consistent in all the conducted experiments.

\section{Experiment Results and Discussion}
The proposed D-LGMD model includes a spatial-temporal filter, which allows discrimination of angular velocity and warns of imminent collision when the output MP exceeds a specified threshold. %实际上select的是image的角速度，而只有即将碰撞的物体恰好容易满足既有较大的扩张轮廓，且轮廓扩张速度容易escape from the rejection band.
Systematic experiments were conducted with the aim of assessing the capability of the D-LGMD model when operating in different visual scenes. We conducted both qualitative comparisons with other LGMD models\footnote{In the following sections, "LGMD model" refers to our previous model used in simple UAV flight~\cite{zhao2018bio}. Otherwise, if it refers to another LGMD model, it will be claimed additionally.} to reflect the success of the proposed synaptic mappings and also quantitative parameter sensitivity experiments to analyse the feature of D-LGMD.
Both simulation and real flight FPV video experiments demonstrate that the proposed D-LGMD model has enhanced selectivity to looming. This is particularly the case when the quadcopter performs agile flight in complex visual scenes.

\subsection{Experimental Setup}
\begin{figure}
	\centering
	\includegraphics[width=0.7\linewidth,height=0.20\textwidth]{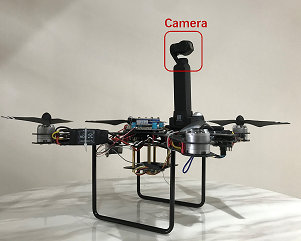}
	\centering
	\caption[]{Quadcopter Platform. The camera is glued on the quadcopter to capture FPV videos.}
	\label{fig:quadcopter1}
\end{figure}

The quadcopter platform used in this study is inherited from our previous research~\cite{zhao2019lgmd}, which is program-controlled to achieve: hover, rotate, accelerate and uniform speed flight tasks. In order to assess different algorithms with identical input image sequences, we used a webcam (OSMO Pocket) , which is fixed on the quadcopter, to record real flight FPV videos. The quadcopter platform is shown in Fig. \ref{fig:quadcopter1}. 
Visual stimuli input to the neural network comprised both simulated objects and real flight FPV videos.
The simulations contain rendered scenes created with Unity Engine software, and basic approaching cubes generated in MATLAB. The neural network was running on a laptop with 2.5GHz Intel Core i5 CPU and 8GB memory.%, after prepared the above mentioned materials and input stimuli.

\subsection{Characteristics of D-LGMD}
\begin{figure}
	\centering
	\subfloat[]
	{
		\includegraphics[width=0.9\linewidth, 
		height=0.14\textheight]{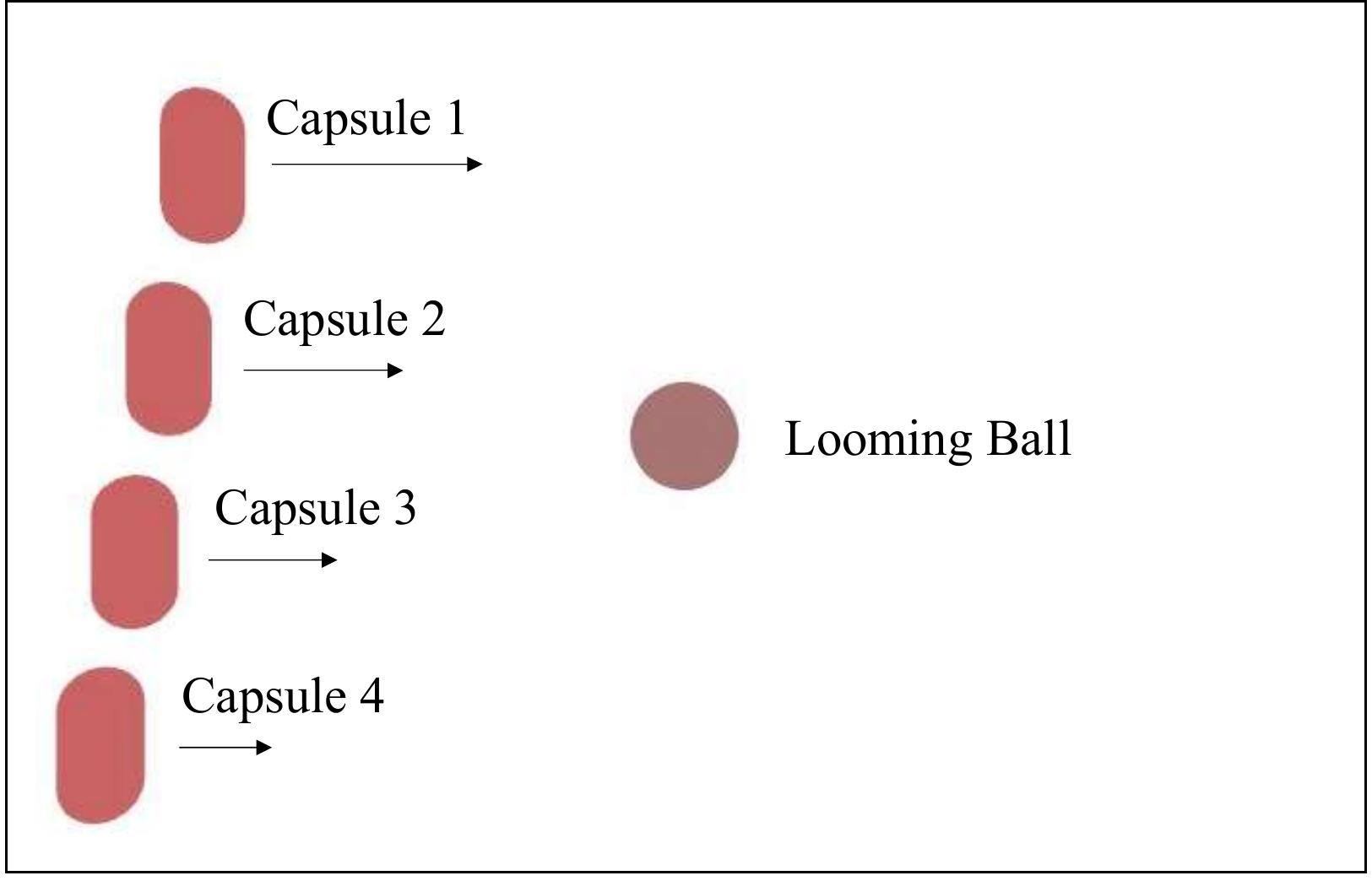}
	}
	\flushleft
	\subfloat[]
	{
		\includegraphics[width=1.0\linewidth, height=0.25\textheight]{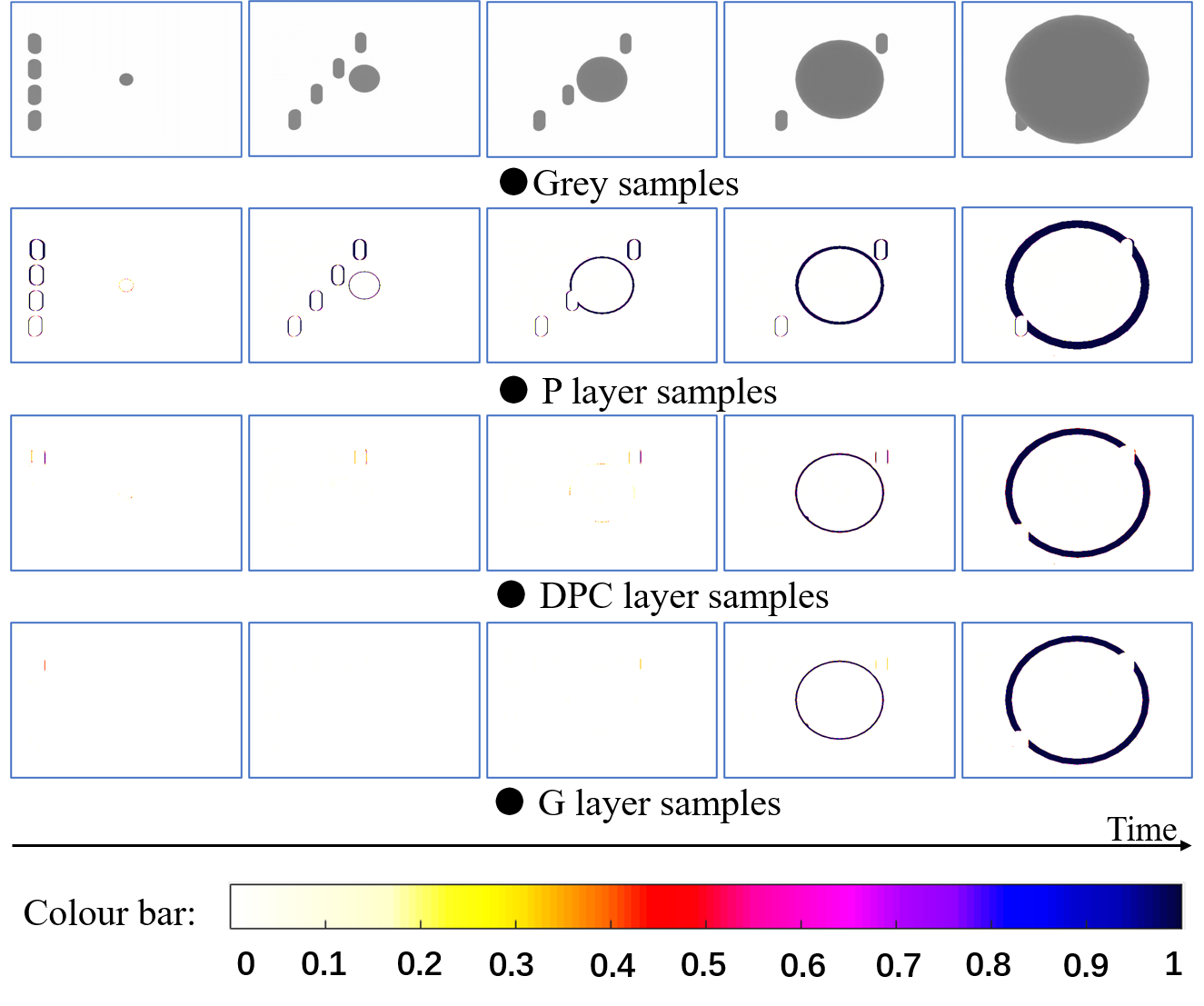}
	}
	\caption{Dissection of each layer's image process with Unity rendered input scene. For better resolution, P layer, S layer and G layer results are transformed to heat-map and presented with colour bar. \textbf{(a)} example input scene. \textbf{(b)} Sampled frames of each layer (at 1, 50, 65, 75 and 80 frame respectively). It is noted that in the heat-map, the intensity tends to be congested to the two extreme ends of the colour bar, as the results of the DPC filter.
	}\label{Fig:Example_Antonomy}
\end{figure}

To intuitively illustrate the layered network's signal processing, an example of unity rendered scene is shown in Fig. \ref{Fig:Example_Antonomy} (a). And a timeline of output data from different layers is shown in Fig. \ref{Fig:Example_Antonomy} (b). The input scene contains a single looming ball with a uniform approaching speed and 4 capsules translating from left to right at different speeds. The P layer acquires luminance changes without any selection, so that stimuli of all the capsules and the looming ball are passed to the DPC layer. The DPC layer discriminates between angular velocities and dramatically attenuates stimuli from translating objects, therefore, it polarised the output into extreme high or low intensity. Subsequently, the G layer further filters sparse or decayed signals and amplifies the grouped excitation under the mediation of FFI. 

\begin{figure}
%	\flushleft
	\centering
	\subfloat[]
	{
		\includegraphics[width=0.9\linewidth,
		height=0.14\textheight]{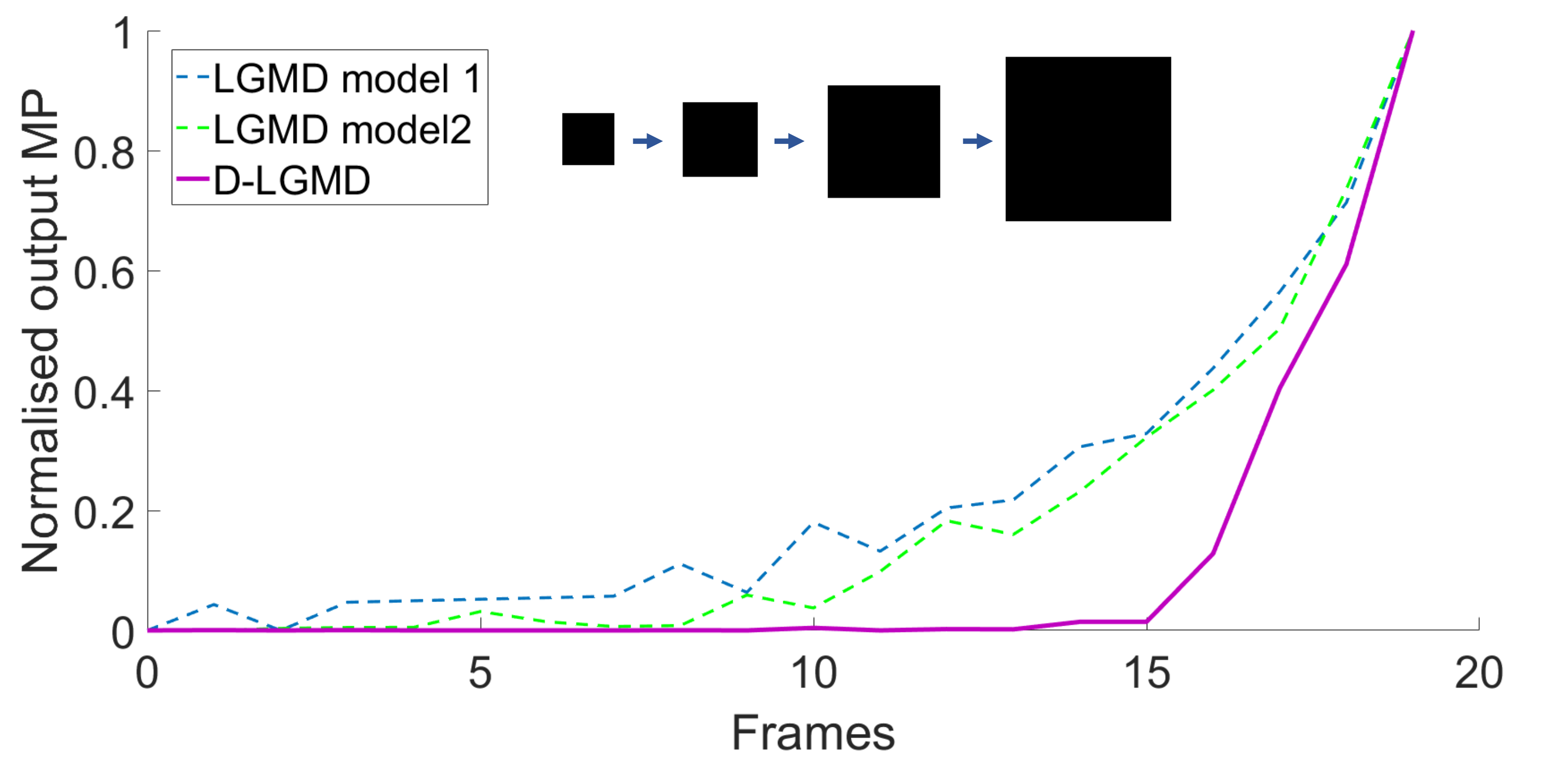}
		\label{fig:Purecube_looming}
	}

	\centering
%	\flushleft
	\subfloat[]
	{
		\includegraphics[width=0.9\linewidth,
		height=0.14\textheight]{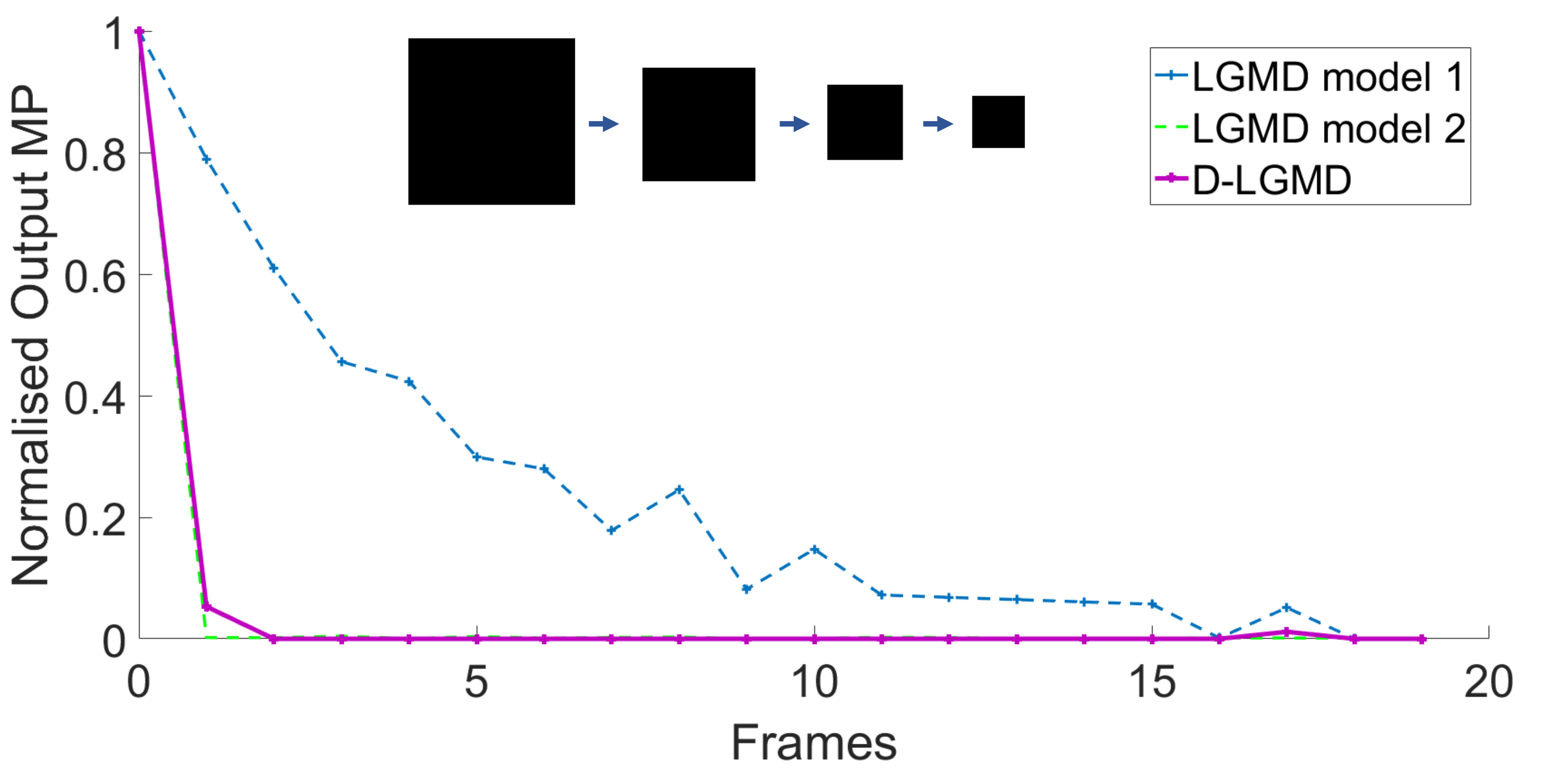}
		\label{fig:Purecube_receding}
	}
	
	\caption{Comparative response for \textbf{(a)} looming and \textbf{(b)} receding object between the LGMD and D-LGMD models. The input stimuli are simulated image sequences of a looming or receding cube (sampled and stamped on the graph).  }\label{Fig:ComparePurecube}
\end{figure}

In a second scenario, simulated stimuli were used to provide input data to compare the responses of the LGMD and D-LGMD models. The normalised output MPs resulting from a looming and receding cube is shown in Fig. \ref{Fig:ComparePurecube} (a) and \ref{Fig:ComparePurecube} (b) respectively. In the case of the looming cube Fig. \ref{Fig:ComparePurecube} (a), the MPs from both the LGMD and D-LGMD models increased in a non-linear fashion as the cube approaches. An increasing in MP of the LGMD model is apparent in the early stages, initially gradual, but increasing continuously in the later stages. In contrast, the MP output from the D-LGMD model remained silent for most of the period but showed a sharp and rapid rise from about frame 15. This is because D-LGMD model is sensitive only to the preferred image angular velocity and not to other visual cues. It is therefore more effective at distinguishing between an object that is closing dangerously, and one that is far away. Similarly, the D-LGMD model also demonstrated a greater capability to reject receding objects, as can be seen from Fig. \ref{Fig:ComparePurecube} (b). Hence, analogous to working characteristics of the real LGMD neuron of a locust when facing receding objects~\cite{Rind2002Motion}, the MP of D-LGMD model dropped to zero soon after the initial activation. In contrast, the LGMD model does not demonstrate the ability to effectively ignore receding objects. This result indicates the proposed D-LGMD model is robust against receding interfering stimuli.

\begin{figure}
	\centering
	\includegraphics[width=0.9\linewidth,
	height=0.16\textheight]{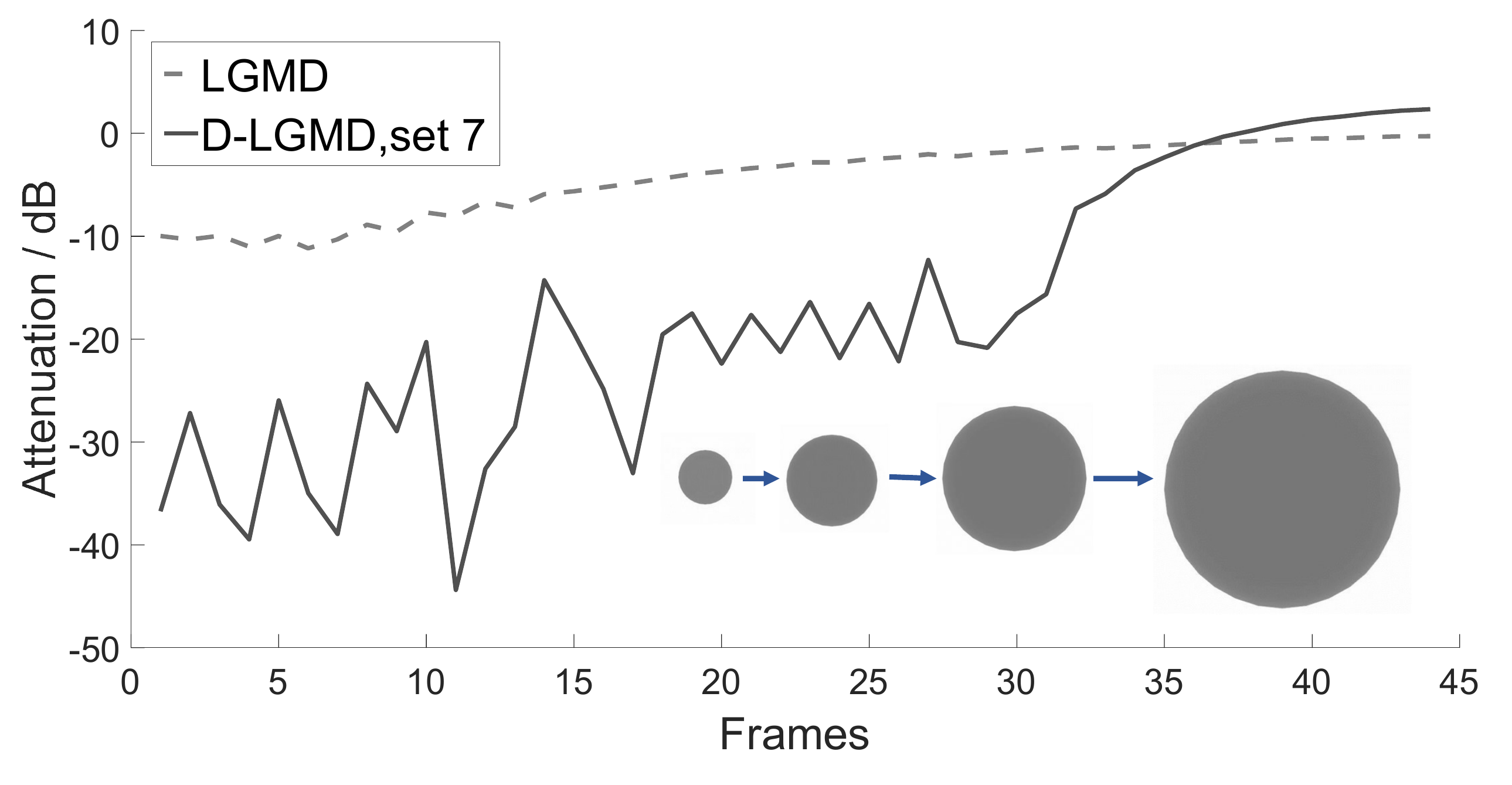}
	\caption{Attenuation change during looming process (D-LGMD parameters: set 7). The input scene is a looming ball generated by Unity engine and presented as grey circles at the bottom right of the figure.}
	\label{fig:attenuationlooming}
\end{figure}

We designed the proposed DPC structure as a spatial-temporal filter with the purpose of selectively attenuating the signal in reaction to the different angular velocities of the images. Here we can define the "attenuation" of the DPC layer as the log-transformed function of the summation of DPC pixels and summation of input luminance changes:
\begin{equation}\label{qt:Attenuation}
Attenuation(t) = 10 \lg(\frac{\iint S(x,y,t)dxdy}{\iint P(x,y,t)dxdy})
\end{equation}
Experiments demonstrate that the attenuation is greatly dependent on the angular velocity of the image. For example, Fig. \ref{fig:attenuationlooming} illustrates the changes in attenuation by the D-LGMD model during a looming process. The attenuation of the LGMD model was moderate when reacting to relatively lower angular speeds as shown in the initial stages, and steadily became less intense (i.e. became less negative) as the object moved closer. In contrast, the attenuation by the D-LGMD model was stronger (i.e. more negative) during the initial period and showed a sharper reduction (i.e. became less intense) near the collision point. This means that the D-LGMD model has a much stronger ability to discriminate between the stimuli since the angular velocity of a looming object always increases in a non-linear way.

%is strong for stimuli with relative lower angular speed at the beginning, and rise up as the object getting closer. Compared to LGMD, D-LGMD showed stronger attenuation in start period and a sharper rise near the collision point, which means D-LGMD has stronger ability to discriminate the stimuli as the angular velocity of a looming object always increases non-linearly.

\subsection{Parameter Sensitivity Analysis}
The response preference of D-LGMD model is determined by the spatial-temporal distribution. In this subsection, we discussed several parameters which are critical to define this preference. These comprise: the inhibition strength $a$, standard deviations of excitation and inhibition distributions $\sigma_E,\ \sigma_I$, and time race coefficients $\alpha,\ \beta,\ \lambda$. Parameters sets used in later experiments are listed in Table.\ref{TB:parameter2} for comparison. Please note, parameter set 1-3 present D-LGMD model without radially distributed latency (when $\alpha=\beta=\lambda=0$, $\tau(x,y)=1.$ ), and they are compared in Fig.~\ref{fig:Compare_ht}

\begin{table}
	\centering
	\caption{Comparative Sets of Crucial Parameters}
	\label{TB:parameter2}
	\begin{tabu}{|[1.5pt]c|c|c|c|[1.5pt]c|c|c|[1.5pt]c|c|[1.5pt]}
		\tabucline[1.5pt]{-} 
		& \multicolumn{3}{c|}{\makecell{Temporal \\Distribution\\ Parameters}} & \multicolumn{3}{c|}{\makecell{Spatial\\ Distribution\\ Parameters}} &\multicolumn{2}{c|[1.5pt]}{\makecell{Other\\ Parameters}}  \\ 
		\tabucline[1.5pt]{-} 
		sets & $\alpha$ & $\beta$ & $\lambda$ & $\sigma_E$ & $\sigma_I$ & a & $T_0$ & r  \\ 
		\tabucline[1.5pt]{-}
		set1 & 0 & 0 & 0 & 0.35 & \textbf{1} & 1.5 & 0.5 & 4   \\
		\hline 
		set2 & 0 & 0 & 0 & 0.35 & \textbf{1.8} & 1.5 & 0.5 & 4 \\ 
		\hline 
		set3 & 0 & 0 & 0 & 0.35 & \textbf{2.5} & 1.5 & 0.5 & 4 \\ 
		\tabucline[1.5pt]{-}
		set4 & \textbf{-0.1} & \textbf{0.5} & \textbf{0.7} & 0.35 & 1 & 1.5 & 0.5 & 4 \\ 
		\hline 
		set5 & -0.1 & 0.5 & 0.7 & 0.35 & \textbf{1.8} & 1.5 & 0.5 & 4 \\\hline 
		set6 & -0.1 & 0.5 & 0.7 & 0.35 & \textbf{2.5} & 1.5 & 0.5 & 4 \\	\tabucline[1.5pt]{-}
		set7 & -0.1 & 0.5 & 0.7 & 1 & 5 & 1.5 & 0.5 & \textbf{4} \\  
		\hline 
		set8 & -0.1 & 0.5 & 0.7 & \textbf{1} & \textbf{5} & 1.5 & 0.5 & 6 \\\hline 
		set9 & -0.1 & 0.5 & 0.7 & \textbf{1.5} & 5 & 1.5 & 0.5 & 6 \\ 
		\tabucline[1.5pt]{-}
		
		%	set10 & \textbf{-0.6} & \textbf{0.4} & \textbf{0.6} & 1 & 5 & 1.5 & 0.5 & 6 \\  
		%	\hline 
		%	set11 & -0.6 & 0.4 & 0.6 & 1 & 5 & \textbf{1.0} & 0.5 & 6 \\  
		%	\hline 	
		%	set12 & \textbf{0} & \textbf{0} & \textbf{0} & 1 & 5 & 1.5 & 0.5 & 6\\  
		%	\hline 
	\end{tabu} 
	\\
	\footnotesize{$*r$ is synaptic distribution calculating radius used to limit the size of the matrix for computing.}\\
\end{table}

\begin{table}
	\centering	
	\caption{Details of Input Image Sequences}
	\label{TB:InputSequences}
	\begin{tabular}{|c|c|c|c|c|}
		\hline 
		\makecell{Image \\ Sequence}   & \makecell{Backgrounds\\ Complexity} & \makecell{Attitude\\ Motion} & \makecell{ Object \\Texture} & \makecell{Collision \\Frame} \\ 
		\hline 
		Group1 & \makecell{Cluttered\\ Indoor} & \makecell{Pitch, \\Accelerating} & \makecell{Pure \\Colour\\ Chair} & 120 \\ 
		\hline 
		Group2 & \makecell{Cluttered\\ Indoor} & \makecell{Pitch, \\Accelerating, \\Decelerating} & \makecell{Gridding \\Pattern} & 140 \\ 
		\hline	
		Group3 & \makecell{Rotation, \\Indoor} & Yaw & Notebook & 135 \\ 
		\hline 
		Group4 & \makecell{Simple\\ Indoor} & \makecell{Take off,\\ Pitch} & \makecell{Gridding\\ Pattern} & 210 \\ 
		\hline 
		Group5 & \makecell{Simple\\ Indoor} & \makecell{Take off,\\ Pitch} & Carton & 196 \\ 
		\hline 
		
		\hline 
	\end{tabular} 
\end{table}

\begin{figure}
	\centering
	\includegraphics[width=0.9\linewidth, height=0.3\textheight]{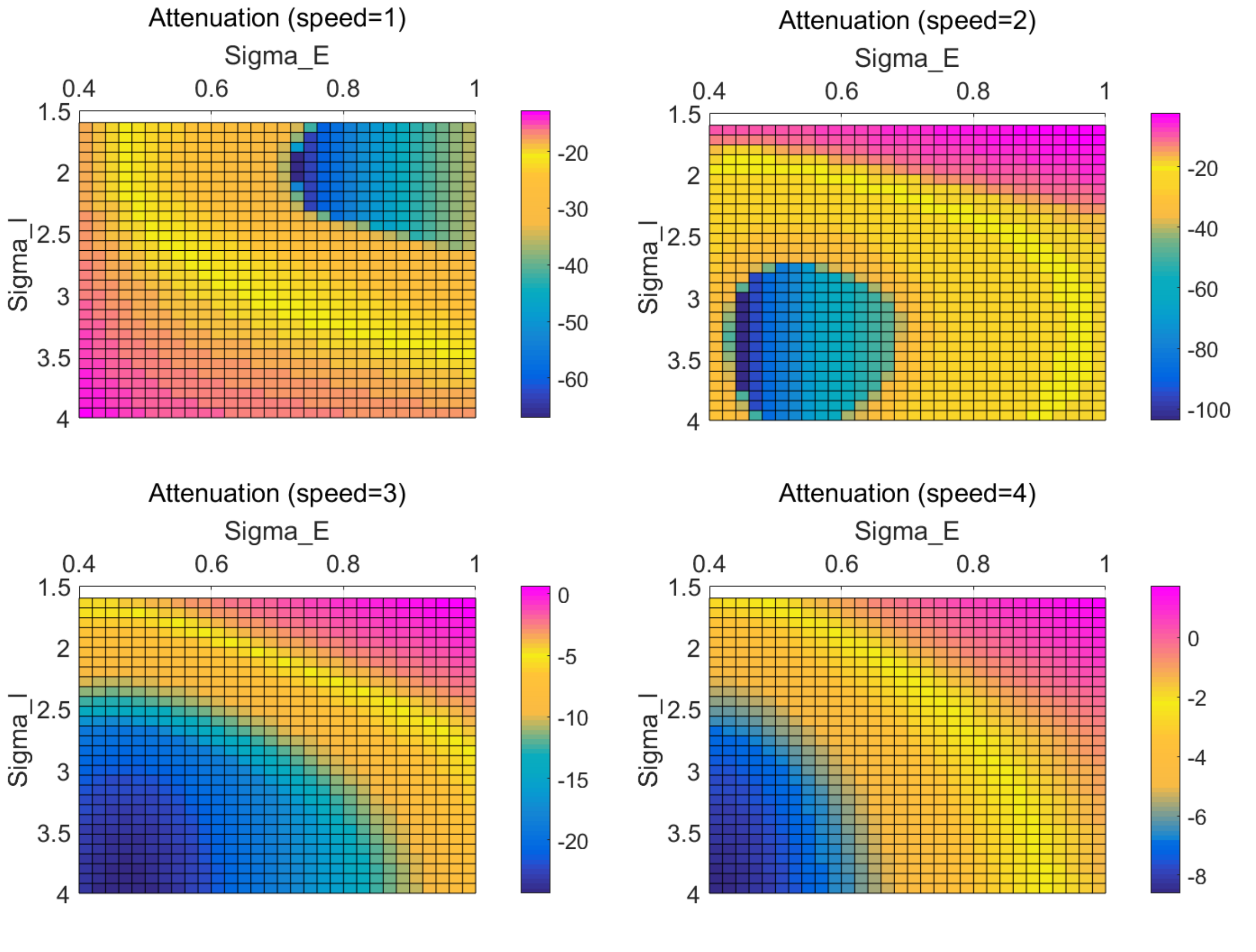}
	\caption{Attenuation analysis in relation to image speed. The input stimuli consist of translating cubes at 4 different velocities: 1,2,3,4 (pixels/frame) which is equivalent to 9,18,27,36 ($^{\circ}/s$) in a $120^{\circ}$, 30 frame rate camera. Note the colour bar in each sub-picture represent different attenuation range. In general, the attenuation is much stronger for objects with relatively lower speed. \textbf{Note: this figure helps to understand the speed selection of the DPC structure. e.g. if the parameters located in the blue region of the top-left sub-figure, considering this region almost does not suppress pixels at speed 2, 3 and 4 in the other sub-figures, this will be a filter that selectively suppress pixels moving at speed 1 pixel / frame.}
		%($\alpha,\beta,\lambda=0$)
	}
	\label{fig:attenuation}
\end{figure}

\begin{figure}
	\centering
	
	\subfloat[]
	{
		\label{fig:Compare_ht}
		\includegraphics[width=0.9\linewidth, height=0.14\textheight]{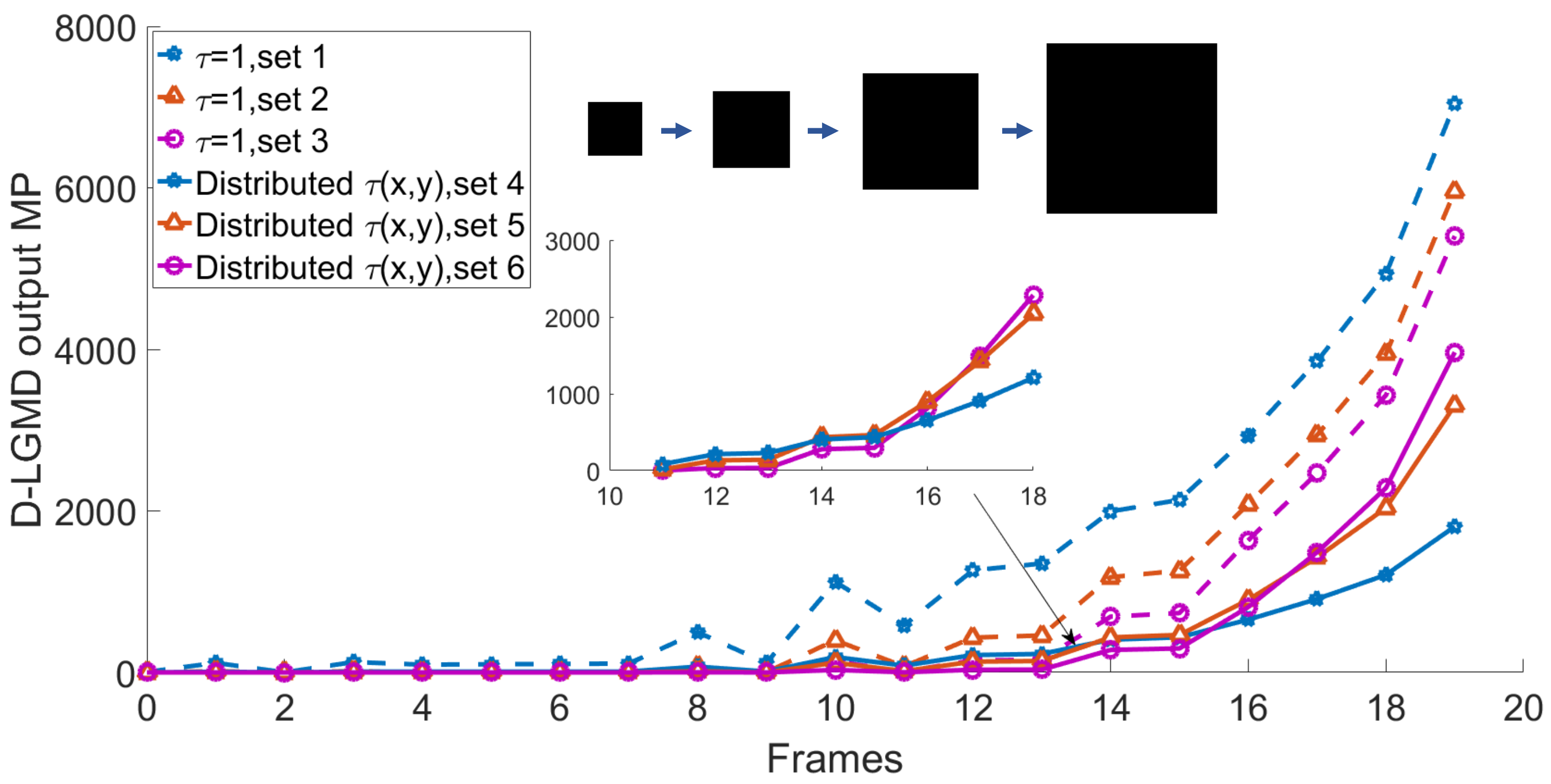}
	}

	\centering
	\subfloat[]
	{
		\label{fig:Compare_ht_Normalised}
		\includegraphics[width=0.9\linewidth, height=0.135\textheight]{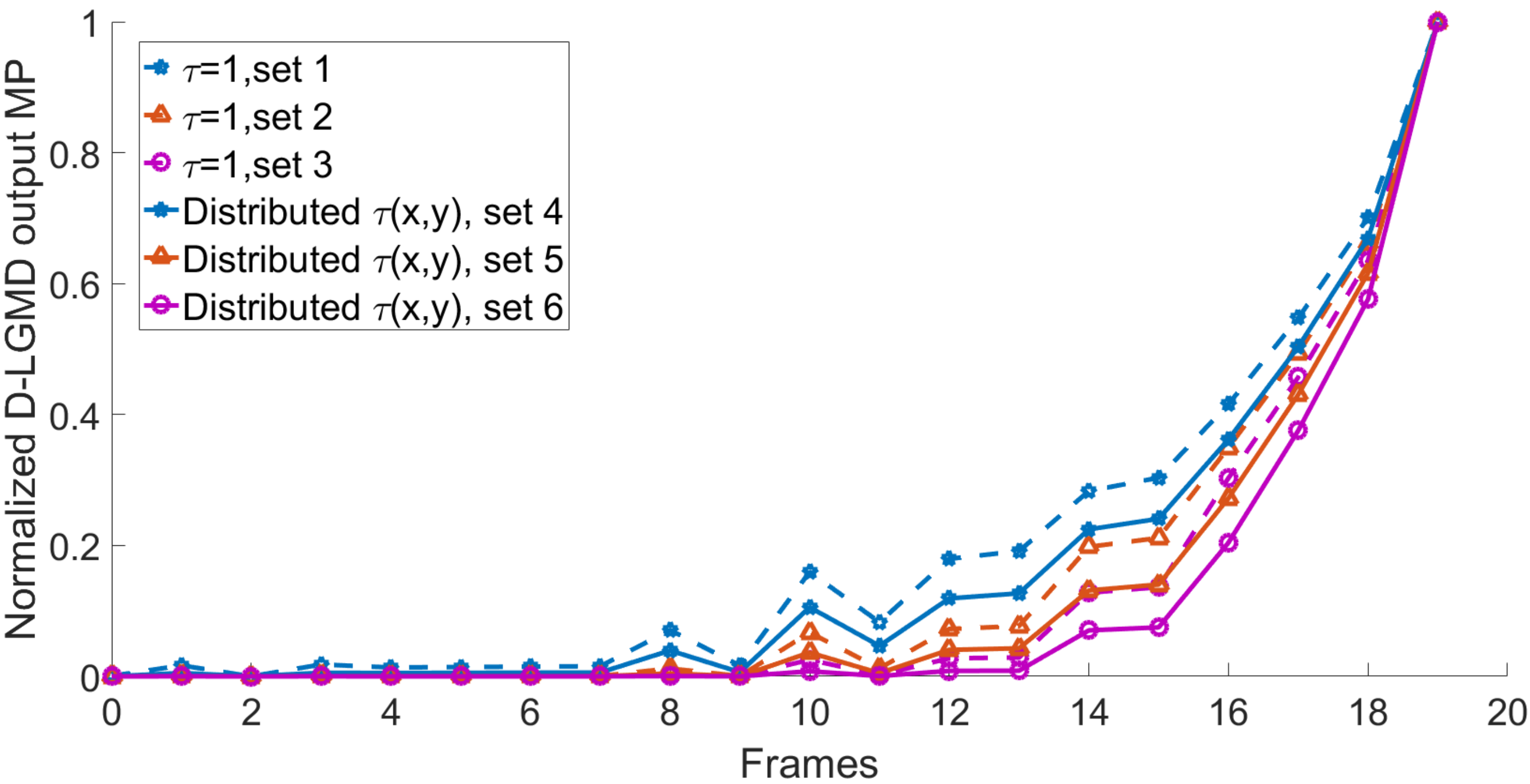}
	}
	\caption{Comparative experiments between distributed time delay and constant time delay. \textbf{(a)} Output MP, \textbf{(b)} Normalized MP. Note: The dashed lines (parameter set 1-3) presented less sharpness than full lines (parameter set 4-6), which indicate that, without the radially distributed latency, the nonlinear preference for fast image motion is reduced.}\label{Fig:Compare_h(t)}
\end{figure}

Fig. \ref{fig:attenuation} shows how changes in $\sigma_E$ and $\sigma_I$ affect the attenuation according to different angular velocities. In general, the D-LGMD model imparts a stronger attenuation (i.e. more negative) for relatively lower speed objects and vice versa. Specifically, increasing $\sigma_I$ enhances the inhibition towards stimuli with higher angular velocities, while increasing $\sigma_E$ reduces the attenuation towards the coherently expanding edges of stimuli with higher angular velocities because they mutually enhance. Thus, tuning the spatial distribution parameters $\sigma_E,\ \sigma_I$ makes it easy to select out preferred angular velocities and to identify image edges of dangerous objects.

Fig. \ref{Fig:Compare_h(t)} demonstrated the advantage of radially extending latency, whose morphological mapping is tuned by time race coefficients $\alpha,\ \beta,\ \lambda$. The radially distributed latency $\tau(x,y)$ sharpened the normalised output curve of both MP and normalised MP. Moreover, in the enlarged details of Fig. \ref{Fig:Compare_h(t)} (a), the parameter sets with weaker response in the beginning climbed over in the looming period. This indicates that as the parameter $\sigma_I$ increases, the performance improves better not only during the looming period (with a stronger output MP) but by showing stronger attenuation when the stimuli is of lower angular velocities. In this case, adjusting $\sigma_I$ for stronger response to stimuli in the looming period does not sacrifice the attenuation towards stimuli that are far away. This occurs because the radially distributed time latency makes it easier for stimuli of higher angular velocities to win the inhibition race as explained previously in Fig. \ref{Fig:Neural_shematic_T}.

\subsection{Performance in UAV FPV Videos}
\label{subsec:FPVVideo}
Finally, the model was challenged with recorded real flight videos. Various input scenes were tested, including a cluttered indoor environment, taking off, multi-axis attitude motion, self-rotation, acceleration and deceleration. Details of the input sequences are listed in Table \ref{TB:InputSequences}.

\begin{figure*}
	\centering
	\subfloat[Grey samples]
	{
		\includegraphics[width=1.0\linewidth, height=0.08\textheight]{Figures/Conundrum_Gray.png}
	}
	\flushleft
	\subfloat[P layer samples]
	{
		\includegraphics[width=1.0\linewidth, height=0.08\textheight]{Figures/Conundrum_Player.png}
	}
	\flushleft
	\subfloat[DPC layer samples]
	{
		\includegraphics[width=1.0\linewidth, height=0.08\textheight]{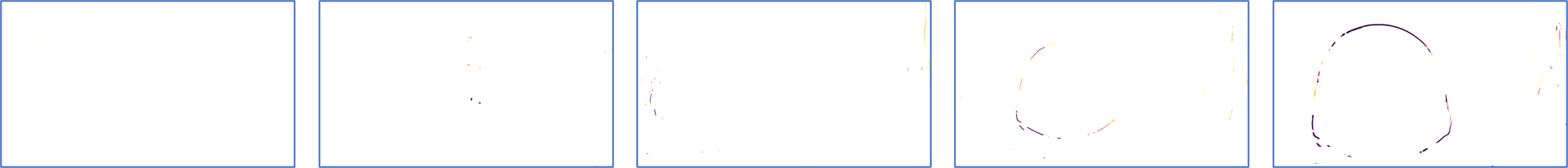}
	}
	\flushleft
	\subfloat[G layer samples]
	{
		\includegraphics[width=1.0\linewidth, height=0.08\textheight]{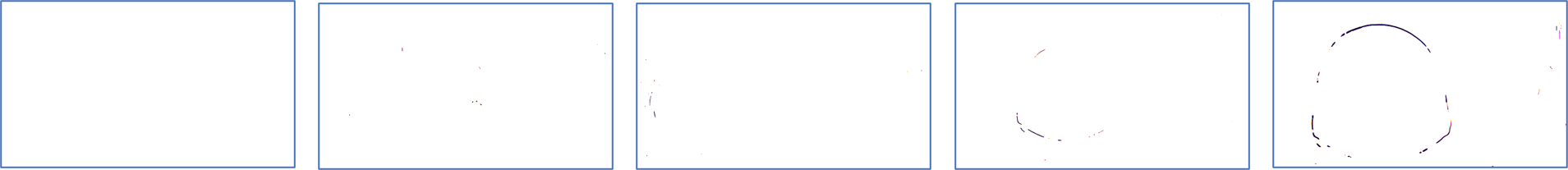}
	}
	\flushleft
	\subfloat[Colourmap]
	{
		\includegraphics[width=1.0\linewidth, height=0.06\textheight]{Figures/colourmap_3}
	}
	%	\flushleft
	%	\subfigure[Output MP ]
	%	{
	%	\label{fig:Conundrum_Compare}
	%	\includegraphics[width=0.45\linewidth, height=0.2\textheight]{Conundrum_Compare.png}
	%	}

	\centering
	\subfloat[Normalised output ]
	{
		\label{fig:Conundrum_Compare_Normalised}
		\includegraphics[width=0.9\linewidth, height=0.3\textheight]{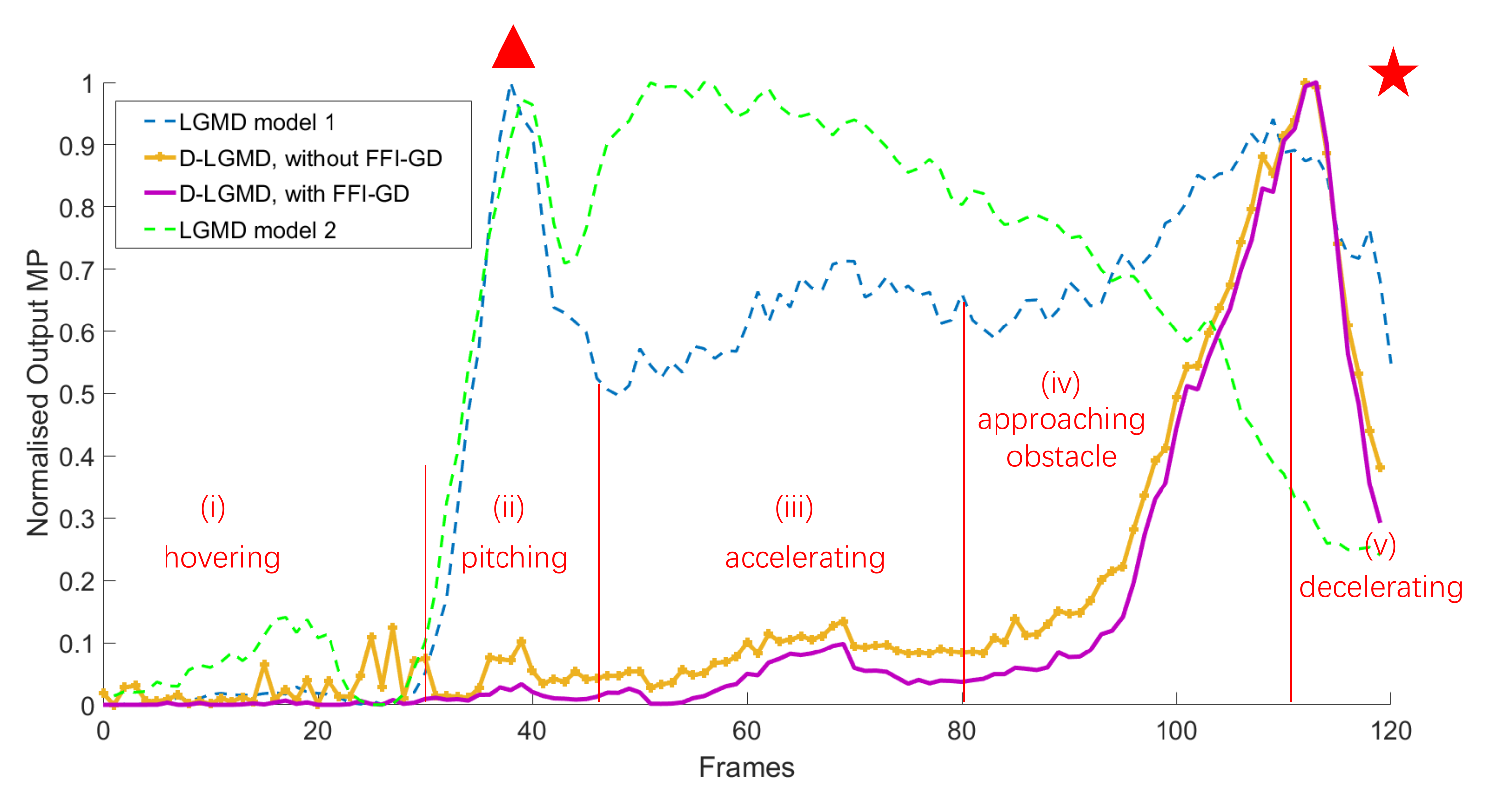}
	}
	
	\caption{Response of 3 LGMD models towards the aforementioned conundrum (input the same scene with Fig. \ref{Fig:Conundrum} (input sequences: Group 2, D-LGMD parameters: set9, example frames are sampled at: 1,35,50,80,100). LGMD model 1: our previous LGMD model applied in simple UAV flight~\cite{zhao2018bio}. LGMD model 2 : the EMD based multiplicative LGMD model by Badia~\cite{Badia2007Blimp}.
	\textbf{Red triangle}: first false positive peak of LGMD result, \textbf{red star}: the collision point. The flight experienced 5 periods as labelled in \textbf{f}: (i) hovering, (ii) pitching, (iii) accelerating, (iv) approaching obstacle(looming), (v) Program controlled decelerating (to avoid hardware damage). Note that the false positive of LGMD model 1 and 2 in periods ii) and iii) is eliminated in the output of D-LGMD model.}\label{Fig:ComplexFlight}
\end{figure*}

In the beginning, the D-LGMD model was challenged by the aforementioned conundrum (the same scene as Fig.~\ref{Fig:Conundrum}) during agile flight (input sequences: Group 2). 
The results are presented in Fig. \ref{Fig:ComplexFlight}. We compared the performance in the same scene of 3 models: our previous LGMD model~\cite{zhao2018bio} (dashed blue curve), the EMD based multiplicative LGMD model~\cite{Badia2007Blimp} (dashed green curve), and the proposed D-LGMD model with/without FFI-GD (yellow and purple curves). 
It is notable that, both dashed curves experienced strong false positives in pitching and accelerating period. While in contrast, the D-LGMD model remains almost silent in these non-collision periods. Instead, as it nears the collision point, the D-LGMD responded a swift activation, and the output MP rises sharply to a high peak. The yellow curve (D-LGMD, without FFI-GD) demonstrates that strong preference for looming emerges after the DPC structure, the performance is satisfactory so that it can even work without post-processing. The purple curve (D-LGMD, with FFI-GD) shows the FFI-GD strategy works well to eliminate small spikes and smooth the curve.

\begin{figure*}
	\centering
	\subfloat[Input example]
	{
		\includegraphics[width=0.9\linewidth, height=0.08\textheight]{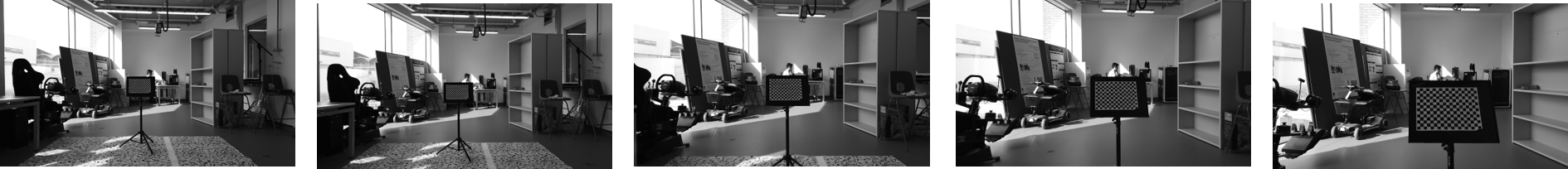}
	}
	
	%	\subfigure{}
	%	{
	%		\includegraphics[width=1\linewidth, ]{Conundrum_Pattern.png}
	%	}
	\centering
	\subfloat[Normalised output]
	{
		\includegraphics[width=0.9\linewidth, height=0.28\textheight ]{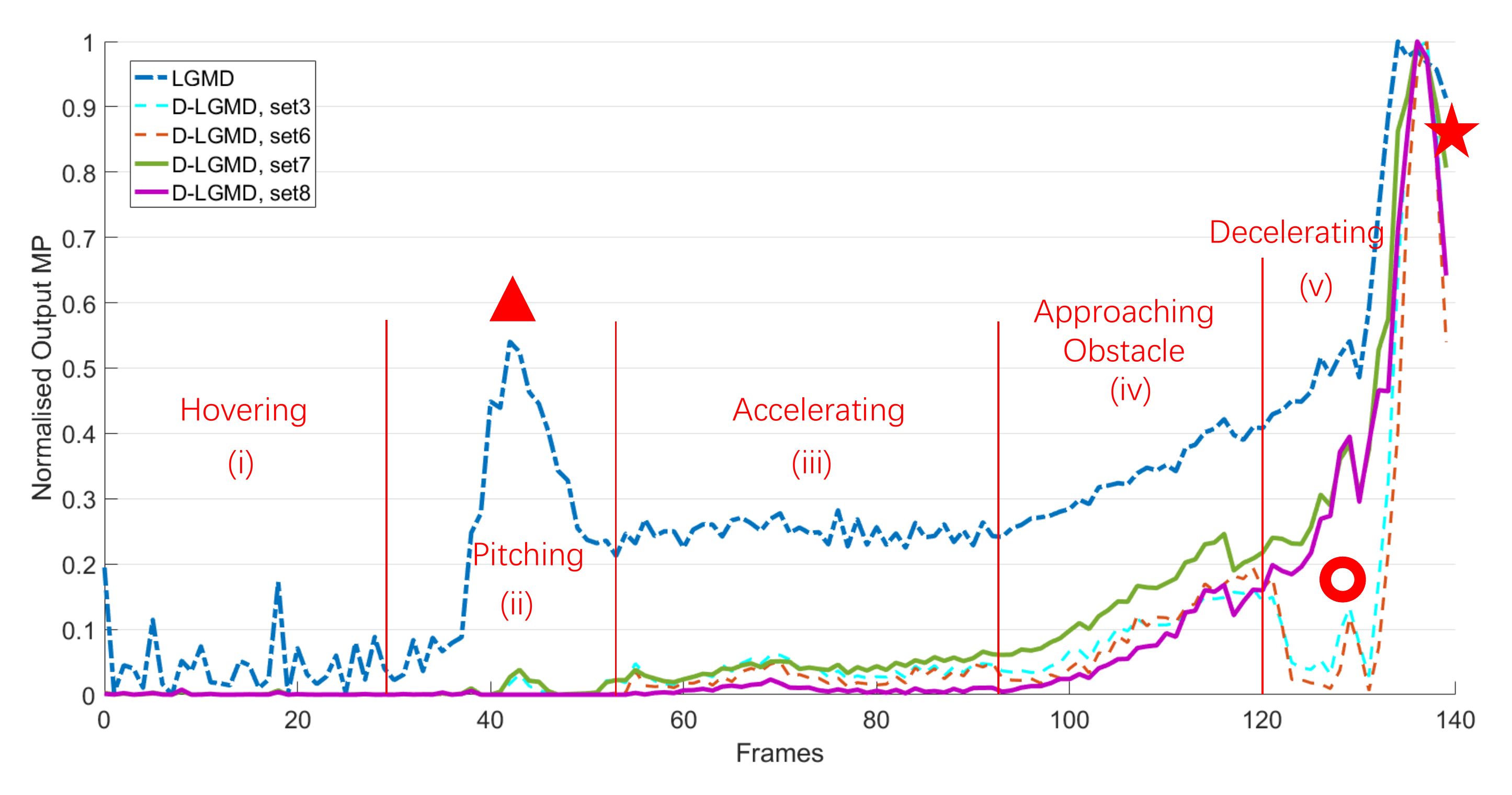}
	}
	\caption{Complex indoor flight (input sequences: Group 2). Collision occurred at frame 140, Pitching and accelerating started at frame 40. Near frame 125 (the red circle), the quadcopter was program controlled to slow down (in order to reduce physical damage in collision). \textbf{Red triangle}: first false positive in LGMD model, \textbf{red circle}: unexpected negative during program controlled deceleration near the collision point, \textbf{red star}: the collision point. Set 3 and set 7 each show a small response near frame 40 (pitching). This small response is eliminated in set 6, 8 when implemented with temporal distribution (set 6) or increased the calculating radius (set 8) respectively. Set 3 and 6 were largely affected by the decelerating process and leading to drop downs near frame 120. This indicate that these two sets (with smaller $\sigma_E$, $\sigma_I$  and calculating radius r) may not have consistent performance when faced with a decelerating looming object. }\label{Fig:Conundrum_Pattern}
\end{figure*}

The model was also tested with the same scene but but with a different obstacle with a different shape and pattern, as shown in Fig. \ref{Fig:Conundrum_Pattern}(input sequences: Group 2). We also evaluated parameter sensitivity during the test (Fig. \ref{Fig:Conundrum_Pattern}, sets 3,6,7\&8). The selected parameter sets were not optimised but chosen as follows: 
\begin{itemize}
	\item Set 3 rarely has excitatory distribution and the temporal latency is constant.
	\item Set 6 involves distributed latency based on set3.
	\item Set 7 has wider excitatory and inhibitory distribution kernel but limits the calculating radius, which limits the size of the matrix in the DPC processing in realistic computing, to 4 ($r=4$ cannot fully reflect the kernel).
	\item Set 8 involves extending the calculating radius to 6 and makes full use of the convolution kernel.
\end{itemize}
   
The results clearly show the different effects of modulating the parameters: Set 3 and 7 exhibit a small response near frame 40 (pitching). This small response is eliminated in set 6 and 8 when adding temporal distribution or increasing the calculating radius respectively. Set 3 and set 6 were largely affected by the decelerating process and led to a decrease in output MP near frame 120. This indicates that these two sets (with smaller $\sigma_E$, $\sigma_I$ and calculating radius r) may not perform consistently when facing obstacles during a deceleration.

\begin{figure}
	\centering
	\subfloat[Input example:]
	{
		\includegraphics[width=0.8\linewidth, height=0.05\textheight]{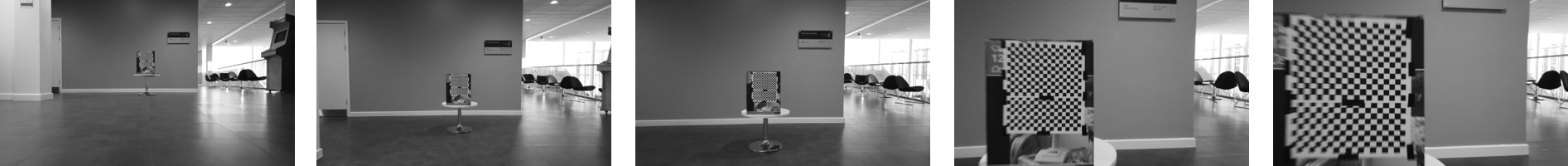}
	}	
	%	\subfigure{}
	%	{
	%		\includegraphics[width=1\linewidth, ]{Indoor_1.png}
	%	}	
	\\
	\subfloat[]
	{
		\includegraphics[width=0.9\linewidth, height=0.13\textheight ]{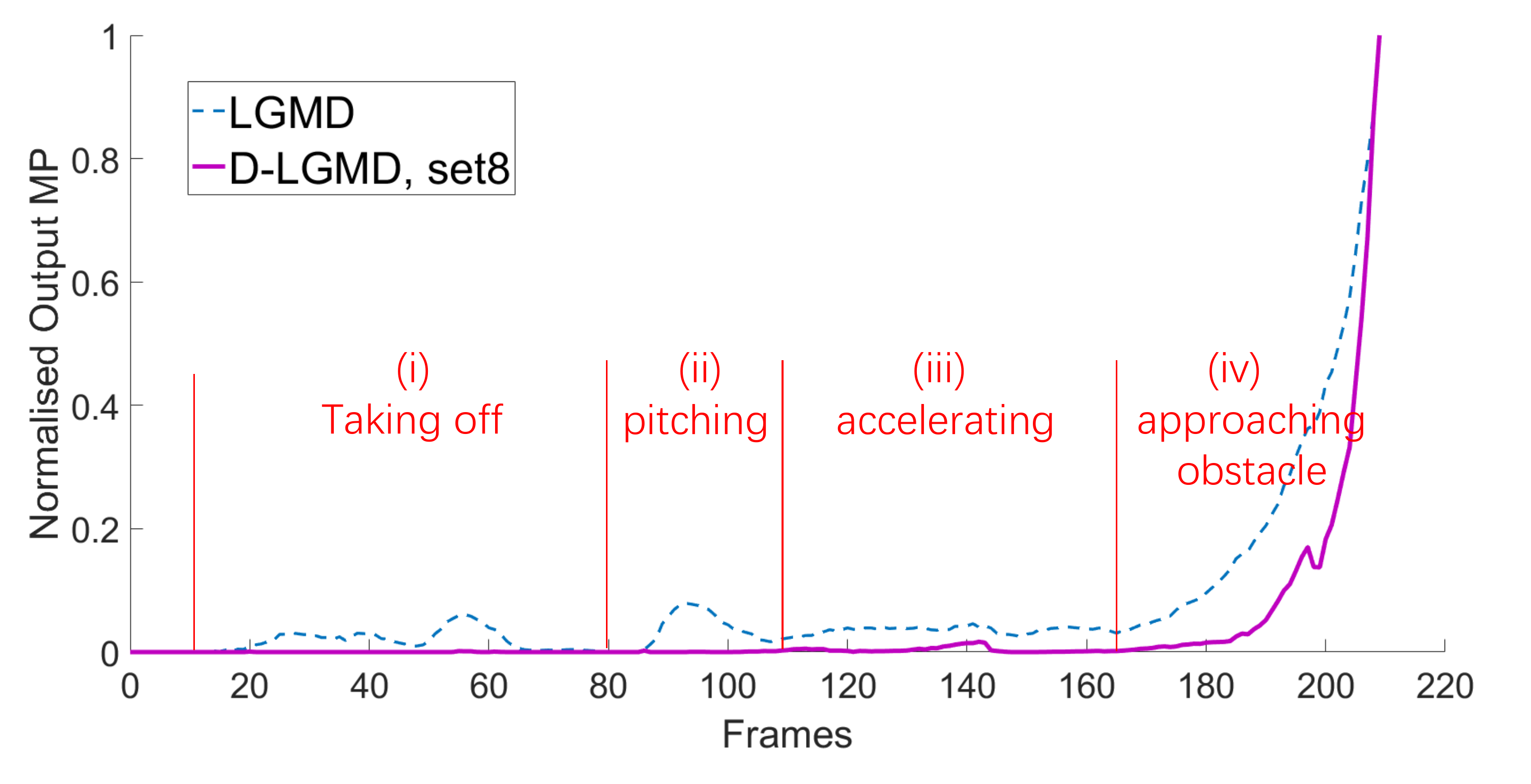}
	}
	\\
	\subfloat[]
	{
		\includegraphics[width=0.9\linewidth, height=0.13\textheight ]{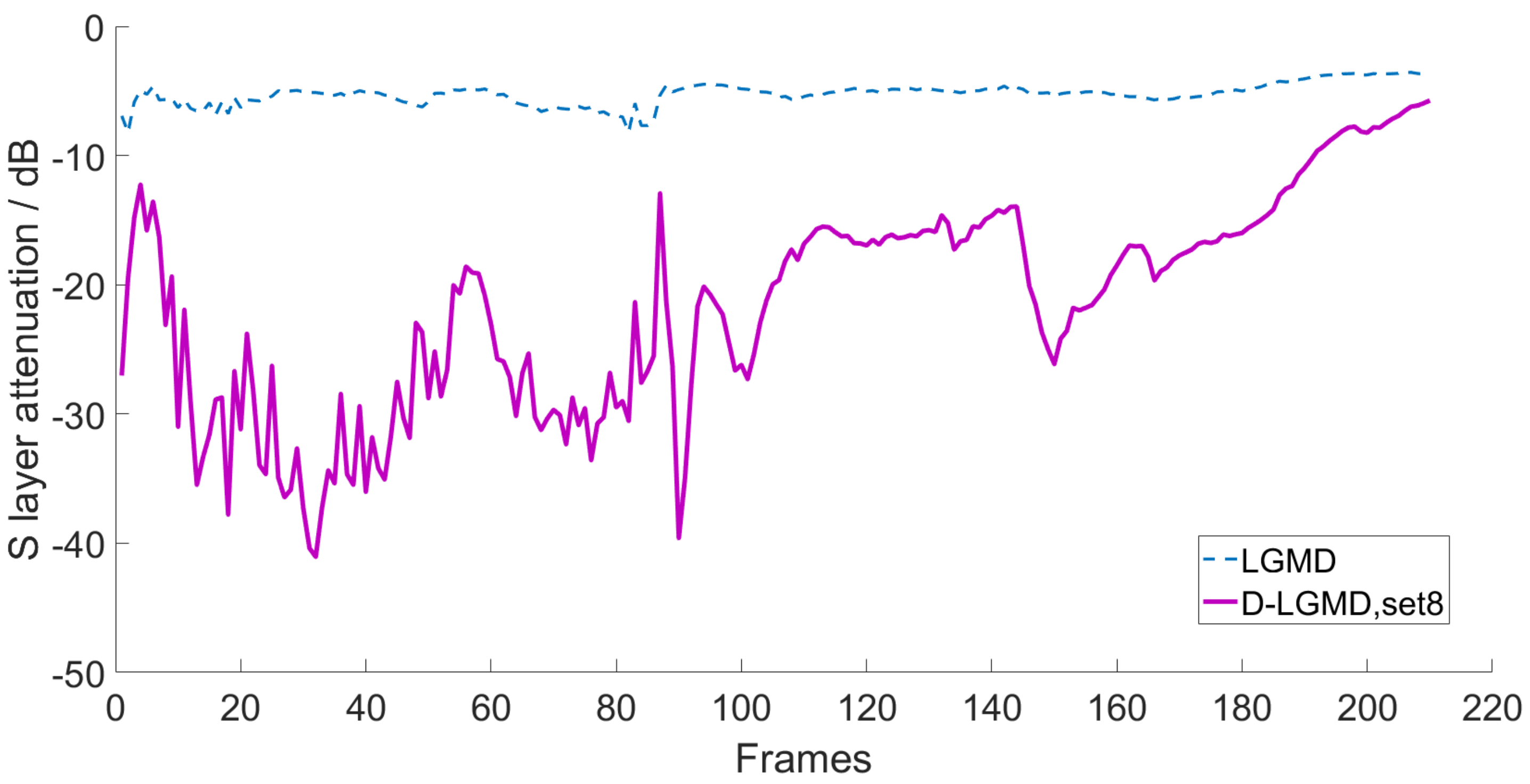}
	}
	\caption{Simple indoor flight (input sequences: Group 4) results of \textbf{(b)} output MP and \textbf{(c)} Attenuation. Collision occurred near frame 210, attitude motion periods are annotated on \textbf{(b)}.}\label{Fig:IndoorSimple_1}
\end{figure}

\begin{figure}
	\centering
	\subfloat[Input example:]
	{
		\includegraphics[width=0.8\linewidth, height=0.05\textheight]{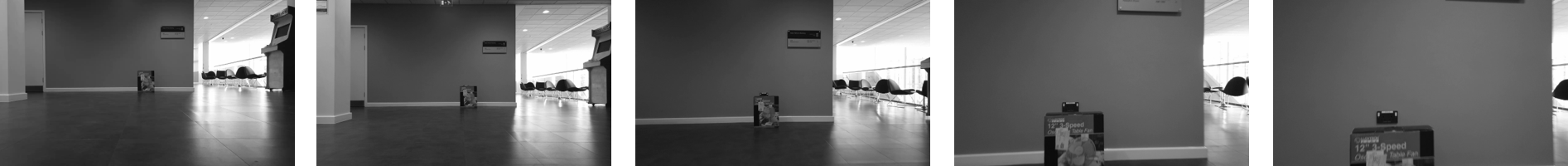}
	}	
	%	\subfigure{}
	%	{
	%		\includegraphics[width=1\linewidth, ]{Indoor_2.png}
	%	}	
	\\
	\subfloat[]
	{
		\includegraphics[width=0.9\linewidth, height=0.13\textheight ]{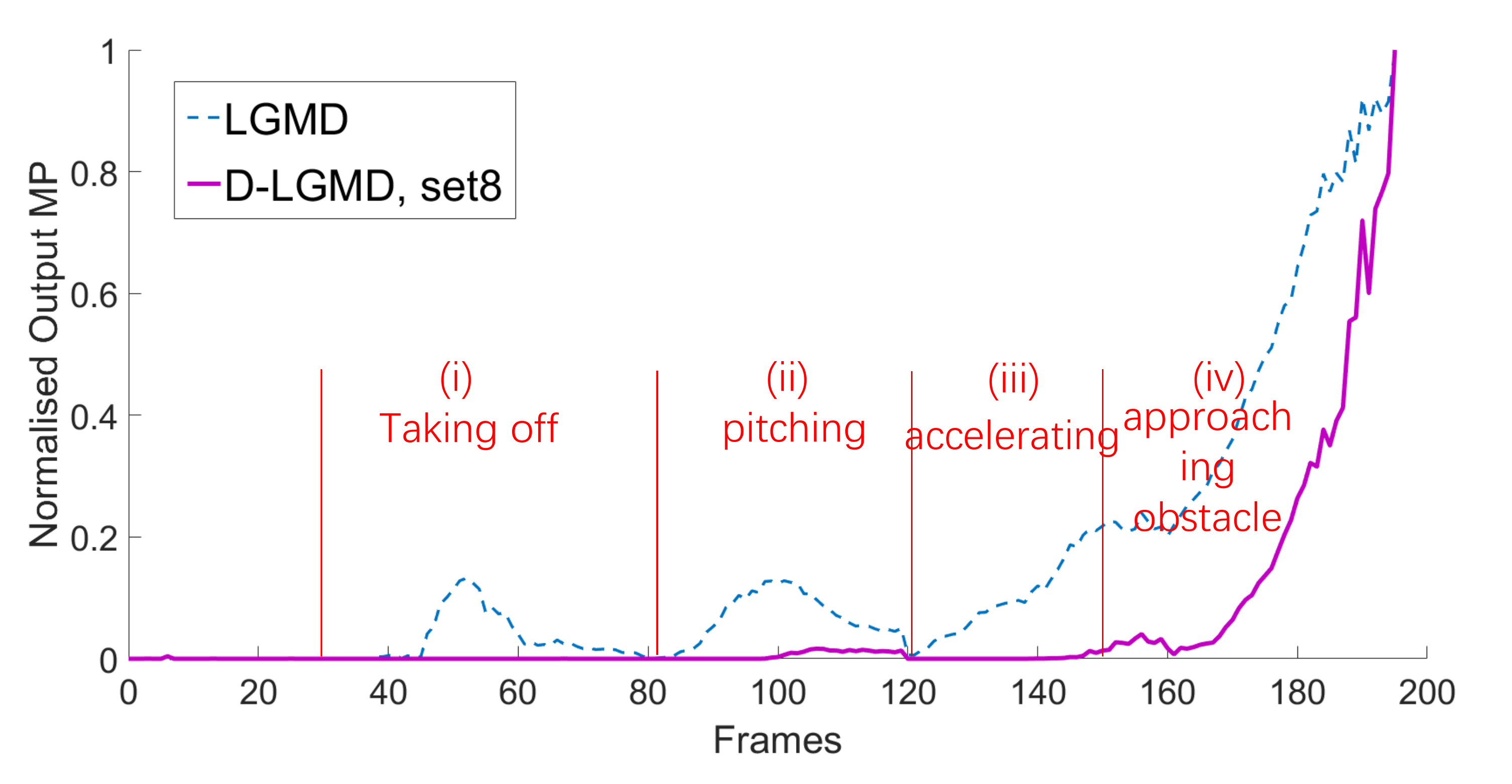}
	}
	\\
	\subfloat[]
	{
		\includegraphics[width=0.9\linewidth, height=0.13\textheight ]{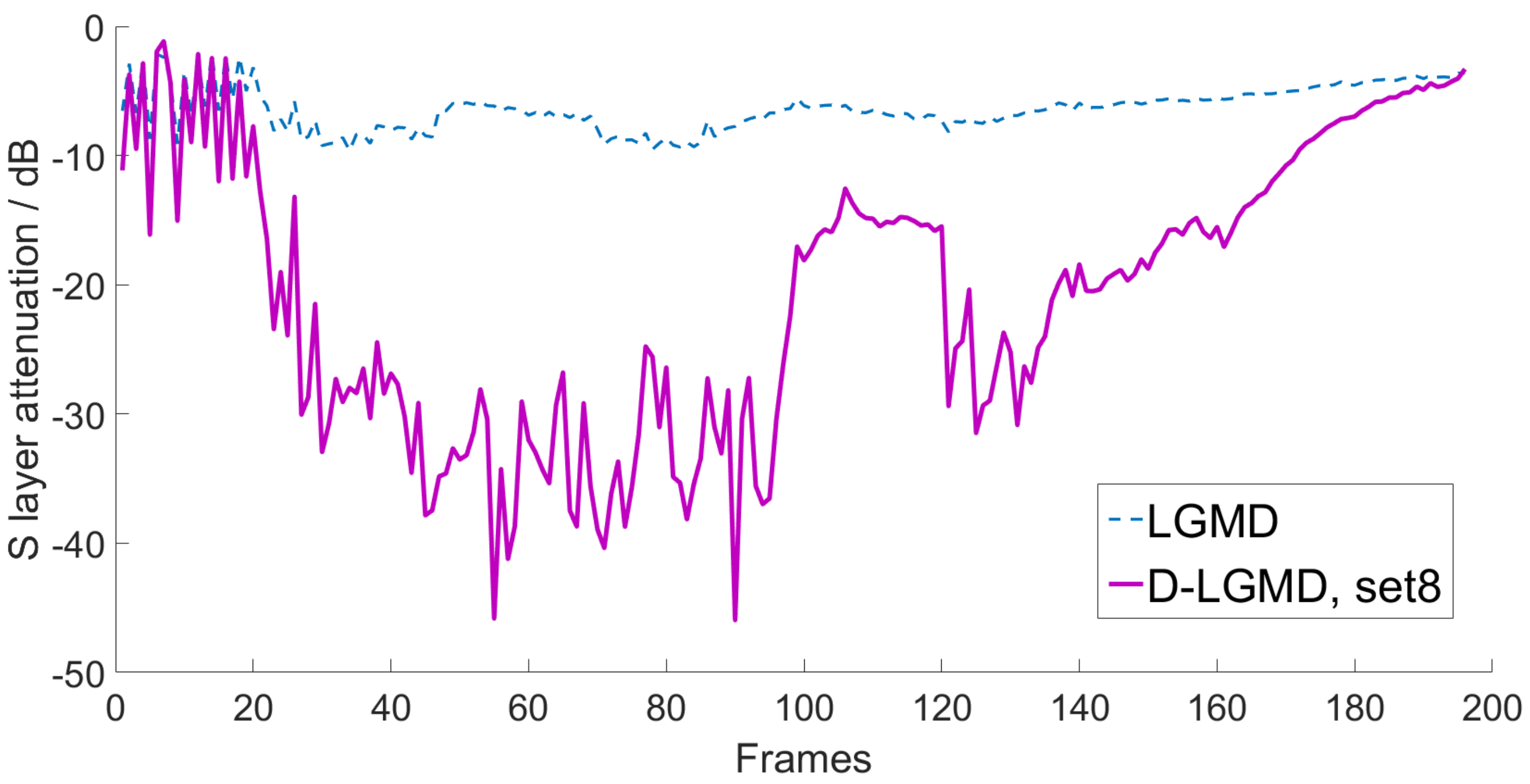}
	}
	\caption{Simple indoor flight-2 (input sequences: Group 5) results of \textbf{(b)} output MP and \textbf{(c)} Attenuation. Collision occurred near frame 198, attitude motion periods are annotated on \textbf{(b)}.}\label{Fig:IndoorSimple_2}
\end{figure}

The D-LGMD also worked well when faced with relatively simple backgrounds (Fig. \ref{Fig:IndoorSimple_1} and Fig. \ref{Fig:IndoorSimple_2}). The results show that both the LGMD and D-LGMD models are able to detect collision in these simple scenes. However, the results from the LGMD show that several small peaks remain during attitude motion. The attenuation curves provide further interpretation: the D-LGMD model showed stronger discrimination in different periods, strong attenuation is observed after taking off, and the attenuation curve prominently rose up during the looming periods. This demonstrate that, compare to the mussy image motions in taking off period, the model prefers the spatial temporal pattern of real looming object.

\begin{figure*}
	\flushleft
	\subfloat[Grey samples]
	{
		\includegraphics[width=1.0\linewidth, height=0.06\textheight]{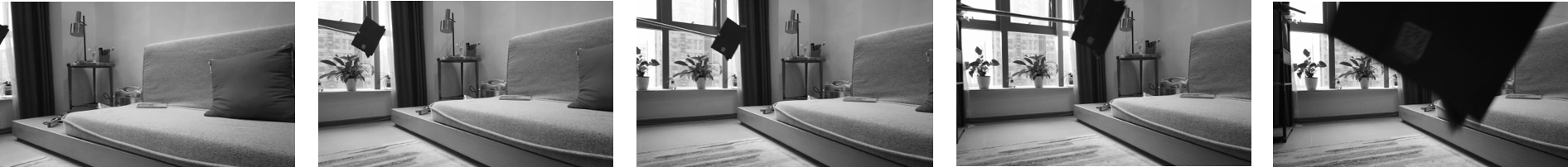}
	}
	\flushleft
	\subfloat[Player Samples]
	{
		\includegraphics[width=1.0\linewidth, height=0.06\textheight]{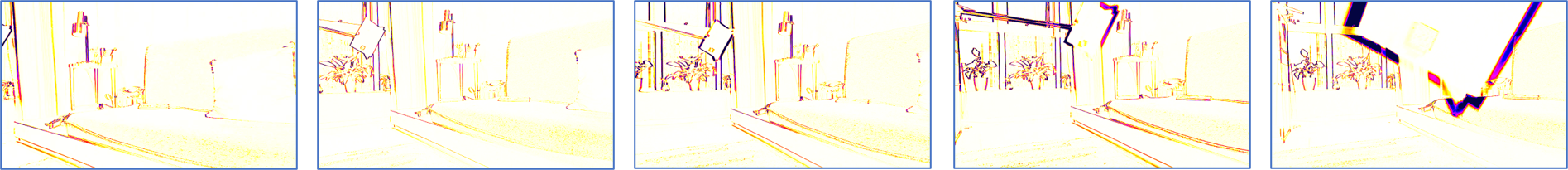}
	}
	\flushleft
	\subfloat[S layer samples]
	{
		\includegraphics[width=1.0\linewidth, height=0.06\textheight]{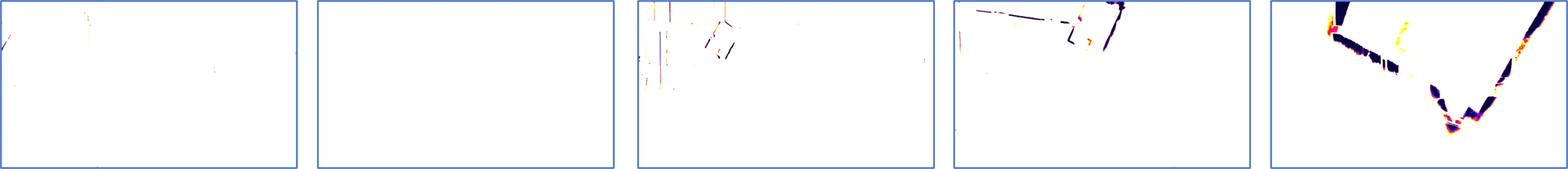}
	}
	\flushleft
	\subfloat[G layer samples]
	{
		\includegraphics[width=1.0\linewidth, height=0.06\textheight]{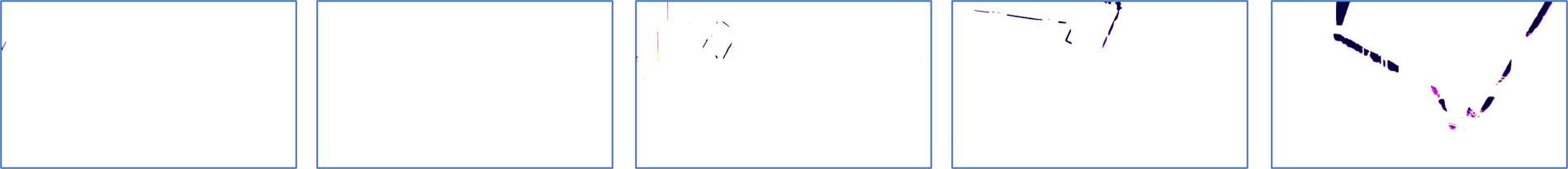}
	}
	\flushleft
	\subfloat[Colourmap]
	{
		\includegraphics[width=0.9\linewidth, 	height=0.06\textheight]{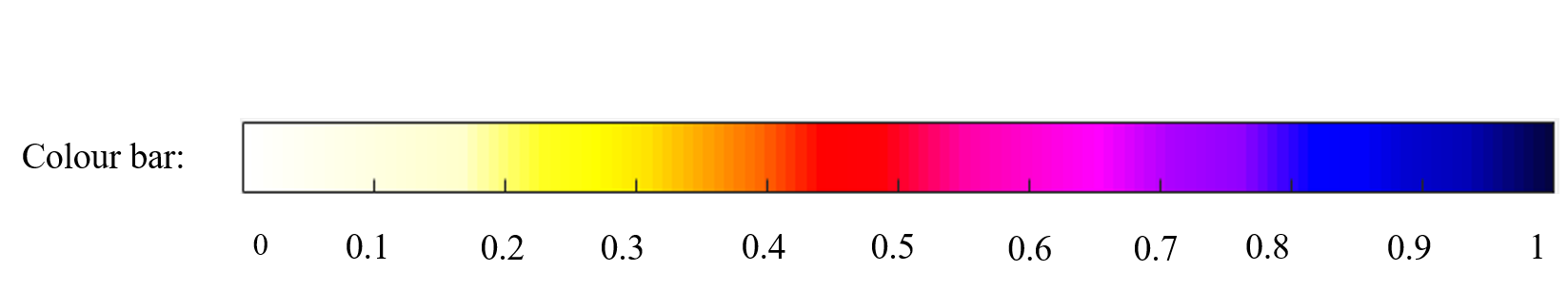}
	}
\\
	\subfloat[]
	{
		\includegraphics[width=0.48\linewidth,height=0.14\textheight]{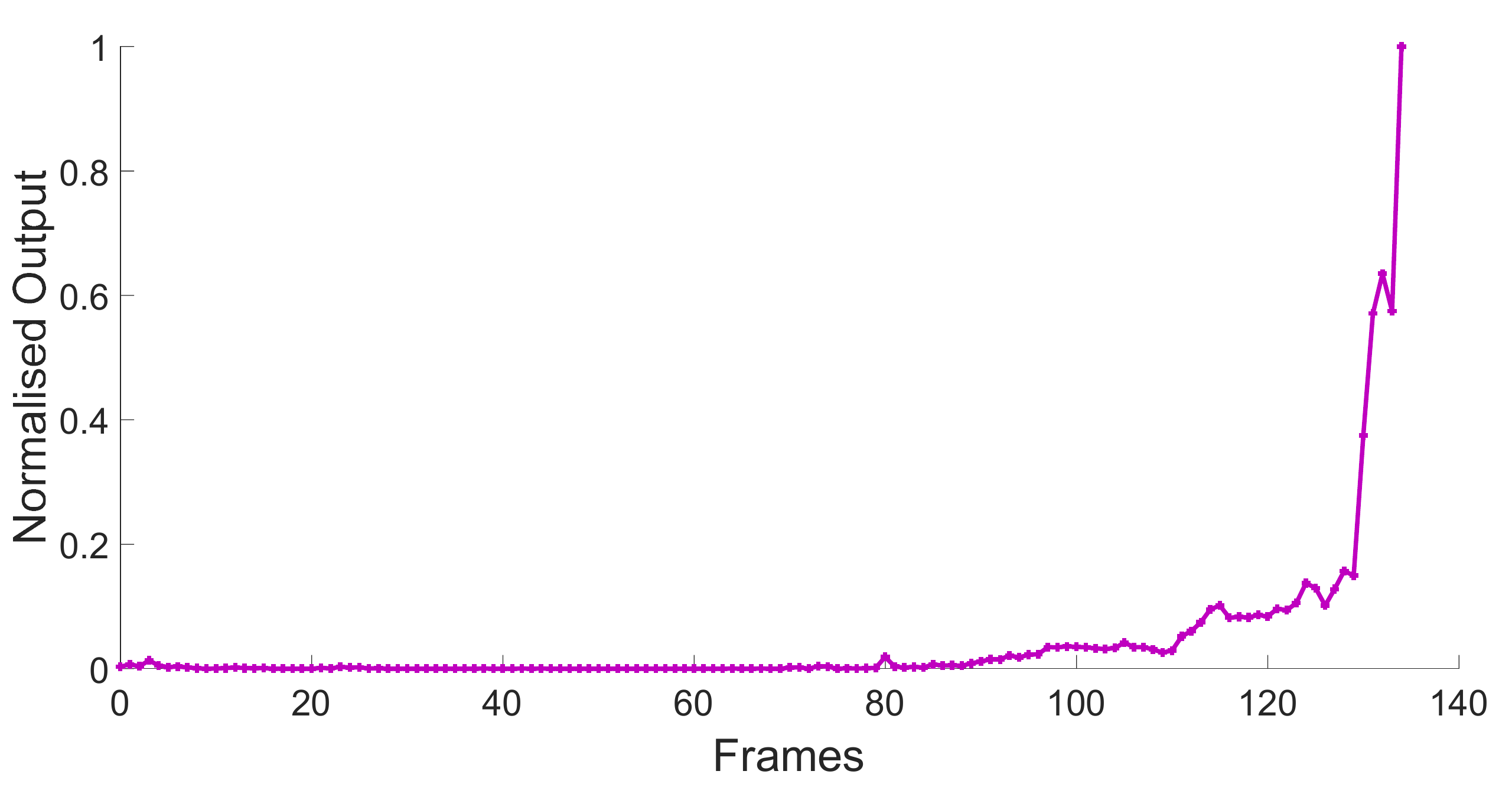}
	}
	\subfloat[]
	{
		\includegraphics[width=0.48\linewidth,height=0.14\textheight]{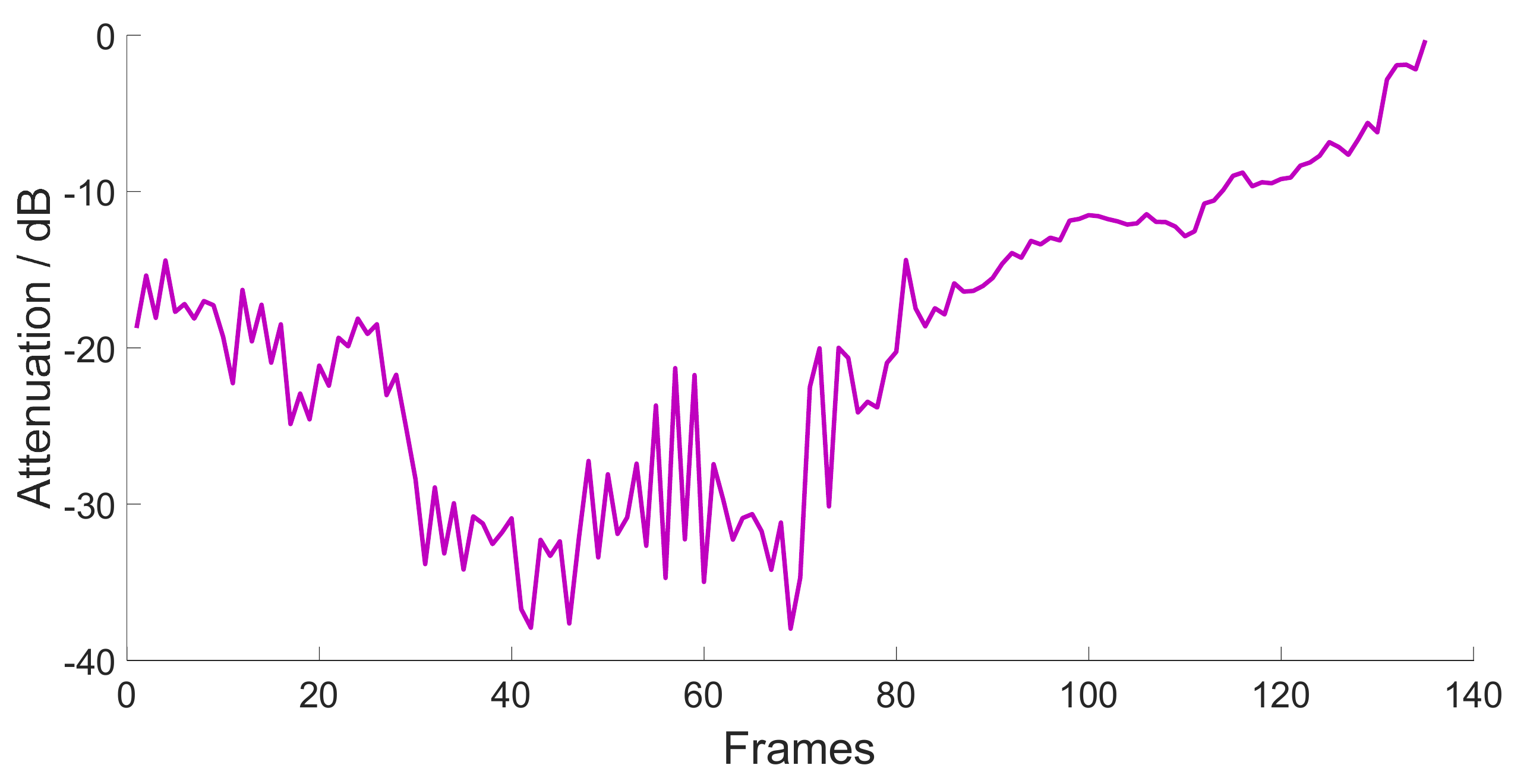}
	}		
	\caption{Detecting Collision during self-rotation (input sequences: Group 3). The quadcopter was program controlled to rotate at ${10 ^\circ}/s$. Example images were sampled at 1,70,100,116,131. It is worth to point out that the attenuation curve showed strong inhibition during rotation and vice versa when the object nears the collision point.}
	\label{fig:rotation}
\end{figure*}

Additionally, we challenged the D-LGMD model during self-rotation (yaw motion). Results in Fig. \ref{fig:rotation} indicate the D-LGMD model preserves the ability to discriminate looming cues during rotational flight.

\subsection{Computation Complexity}
The computational complexity of the proposed DPC layer is mainly determined by the 2D convolutions of the input image sequences with $W_E$ and $W_I$ (equation (\ref{qt:2}) and (\ref{qt:3})),
%As the distribution pattern is pre-determined
which can be implemented in $O(2r^2mn)$ times for an $m\times n$ input image and $r\times r$ size kernel. In other words, the computational complexity is mainly determined by the calculating radius and input image size. %Obviously, c
\textbf[calculate a formulation, including the latency with no latency comparison.]
The calculating radius should cover a significant area of the kernel for its character to be established. %As discussed above, filtering faster background noise requires a "larger" kernel but also leads to increasing computational complexity. The good news is, r
Reducing input image size (provided that the image preserves important features of the looming process) would increase the ability to recognise a looming object because the D-LGMD discriminates between image velocities by pixel interconnections and resizing the input image size also redefines the kernel's impact area corresponding to the real scene. Therefore, the D-LGMD model can work with extremely low-resolution input because reducing the image size makes the kernel cover a larger area and enhances the barrier to background noise. 
We systematically analysed the relationship between the ability to recognise (distinguish-ability) to looming, the calculating radius, input image size and computational complexity by 200 trials running on PC. The results are listed in Table \ref{TB:Computation_Complexity}. The distinguish-ability (DA) is quantified as the average output MP near the peak apex (5 frames before and after) divided by average MP at a false positive point (i.e. Fig.~\ref{Fig:ComplexFlight} (f)): 

$DA=\ average\ (MP_{peak})\  / \   average\ (MP_{false-positive})$. 
Theoretically, when  $DA \leq 1$, it is impossible to select out looming cues from dynamic backgrounds during agile flight. From experimental experience, if $DA > 10$, the model is very competent at filtering out the interfering stimuli and is foreseen to be robust for a range of scenes. As the input size ranging from default to one tenth of the area, the DA initially increased and then decreased. Using parameter set 7 (except r), for all the calculating radii, the best DA results existed when input images were resized to $0.25 \times$ default area ($480\times270$ pixels). The proposed model showed satisfactory DA results even at extremely low resolution ($38\times22$ pixels). An insufficient DA was generated only when computing the default size (1080P) input with $r = 2$, resulting in a $DA=3.87$. This would mean that the looming object would be distinguishable but not sufficiently prominent.

\begin{table*}[ht]
	\centering
	\caption{Computation Complexity}
	\label{TB:Computation_Complexity}
	\begin{tabu}{|[1.5pt]c|c|c|c|c|[1.5pt]}
		\tabucline[1.5pt]{-} 
		Calculating Radius (r) & Image resizing & Distinguish-ability (DA) & O (DPC) & PC Run Time (10 times average) \\ 
		\tabucline[1.5pt]{-} 
		6 & default & 249.33 & $2 \times 6^2 \times 1920\times1080$ & 9.18s \\ 
		\hline 
		6 & 0.5 & 9789.28 & $2 \times 6^2 \times 960\times540$ & 2.75s \\ 
		\hline 
		6 & 0.25 & $>10000$ & $2 \times 6^2 \times 480\times270$ & 1.06s \\ 
		\hline 
		6 & 0.1 & $>10000$ & $2 \times 6^2 \times 192\times108$ & 0.75s \\ 
		\hline 
		6 & 0.02 & 1354.10 & $2 \times 6^2 \times 38\times22$ & 0.60s \\ 
		\tabucline[1.5pt]{-} 
		4 & default & 33.44 & $2 \times 4^2 \times 1920\times1080$ & 8.71s \\ 
		\hline 
		4 & 0.5 & 691.20 & $2 \times 4^2 \times 960\times540$ & 2.54s \\ 
		\hline 
		4 & 0.25 & 4579.40 & $2 \times 4^2 \times 480\times270$ & 0.98s \\ 
		\hline 
		4 & 0.1 & 105.33 & $2 \times 4^2 \times 192\times108$ & 0.70s \\ 
		\hline 
		4 & 0.02 & 184.54 & $2 \times 4^2 \times 38\times22$ & 0.59s \\ 
		\tabucline[1.5pt]{-} 
		3 & default & 9.1 & $2 \times 3^2 \times 1920\times1080$ & 8.70s \\ 
		\hline 
		3 & 0.5 & 96.75 & $2 \times 3^2 \times 960\times540$ & 2.53s \\ 
		\hline 
		3 & 0.25 & 1227.60 & $2 \times 3^2 \times 480\times270$ & 0.97s \\ 
		\hline 
		3 & 0.1 & 36.13 & $2 \times 3^2 \times 192\times108$ & 0.67s \\ 
		\hline 
		3 & 0.02 & 76.06 &  $2 \times 3^2 \times 38\times22$ & 0.57s \\ 
		\tabucline[1.5pt]{-} 
		2 & default & 3.87 & $2 \times 2^2 \times 1920\times1080$ & 8.67s \\ 
		\hline 
		2 & 0.5 & 11.09 & $2 \times 2^2 \times 960\times540$ & 2.49s \\ 
		\hline 
		2 & 0.25 & 32.29 & $2 \times 2^2 \times 480\times270$ & 0.95s \\ 
		\hline 
		2 & 0.1 & 11.76 & $2 \times 2^2 \times 192\times108$ & 0.66s \\ 
		\hline 
		2 & 0.02 & 18.47 &  $2 \times 2^2 \times 38\times22$ & 0.57s \\ 
		\tabucline[1.5pt]{-} 
	\end{tabu} 
	\\
	\footnotesize{Note: DA is defined as the average output MP at peak point divided by average MP at false positive point: \\
	DA $=\ average\ (MP_{peak})\  / \   average\ (MP_{false-positive})$. The parameters used are consistent with set 7 in TABLE \ref{TB:parameter2} (except r). \\The input scene is Group1 in Table \ref{TB:InputSequences}.
	The PC Run Time covers the whole process of loading 120 frames of input images and running the model.}
	%讲一下理论上和经验上，D等于多少的时候，大概是个什么效果。 
\end{table*}

\subsection{Discussion and Future Direction}
As shown and discussed in the previous sections, the proposed D-LGMD model have been verified systematically via the experiments both qualitatively and quantitatively. The qualitative results shown in Fig. 8-10 indicate the D-LGMD model is excellent in discriminating image motion caused by looming objects from that of receding or translating ones. The capability of the DPC layer in filtering image motion based on its angular velocity has been explained in Fig.8, and further demonstrated in the  supplementary material video 1. The quantitative analysis shown in  Fig. 11 and Fig. 12 reflect the characteristic preference on image angular velocity is tunable in our model. More specifically, Fig. 11 also reveals how to tune this model to filter different angular velocity for different scenarios. It is also shown in Fig.12 that a constant temporal latency will lead to less nonlinear selectivity to looming as compared with radially extending latency.

Experiment results in flying scenarios have been shown in Fig. 13 to Fig.17 which demonstrated that D-LGMD is excellent to cope with agile flights in different scenarios. For example, D-LGMD (with/without FFI-GD) has been compared with other two models, where D-LGMD is robust in the pitching and accelerating periods of an agile flight. The proposed D-LGMD model with different spatial-temporal mapping (set3-8) are compared in Fig.14 - all of them are performed excellently compared with the previous LGMD. It is interesting to note that unexpected output drops (indicated by a red circle) appeared just before collision detected. These drops on the other hands demonstrated that the D-LGMD's robustness is very much depending on a minimal synaptic distribution calculating radius $r$, which should possess the characters of the spatial-temporal mappings adequately. The attenuation analysis can help to compare filter efficiencies in signal processing. The attenuation analysis on DPC layer shown in Fig. 15 and Fig. 16 further reveals that the proposed D-LGMD can significantly suppress irrelevant input image motions at different agile flying periods, particularly in the taking off, pitching and accelerating periods.

Recent neural physiological studies have discovered new characteristics of LGMD neuon in locusts to be considered in the model. For example, Zhu~\cite{zhu2018pre} suggested that the distributed excitation increases in response to coherently  expanding edges, FC. Rind~\cite{rind2016two} reveals the new evidences on lateral inhibition, these all suggest the retinotopic connections from photoreceptoers to the LGMD neuron could be more complex than currently modelled. It is also noticed that ON/OFF seperated LGMD models have also shown their ability in replicating looming detecting capacity~\cite{fu2019TCYB}. The performance of an ON/OFF channel separated model with the spatial-temporal distributed synaptic mappings is also a research topic to be investigated in the future.    

%the retinotopic projections from the  photoreceptors  and  interact  with  neighbouring  synapses before they converge

%In future researches, we should think about the mechanism for identifying a looming object to improve the model. As we see, the DPC layer is a filter target to image angular velocity and the final output MP integrates the activation of the whole field of view, Therefore, we can conclude that the D-LGMD model identify a looming object by its angular velocity and angular size on the image. Although the successful performance of D-LGMD in complex dynamic scenarios indicates these two features are important for collision detection, there are other monocular cues may also be significant for it, such as "image expansion".
%In Fig. 9, it can be seen that LGMD model 2 has advantage in distinguishing between approaching and receding.
%Since how does insects integrate monocular visual cues for collision detection is still an open topic, in further researches, more attempts should be conducted to explore how to involve other critical information but with least computational power. We hope to see what chemical reaction will happen when D-LGMD model is combined with other methods.

\section{Conclusion}
%Inspired by state-of-the-art researches of locust's LGMD neuron,
This paper proposed a computational presynaptic neural network model as a solution for collision detection in agile UAV flight applications. Agile flight of a UAV brings ego-motion of the camera which leads to confusing false positive in visual motion based collision detection algorithms. Our solution is to target the neural filter on nonlinear image angular velocity of the looming objects. This is achieved by integrating a series of locally distributed synaptic mappings into the second stage of LGMD process (in the DPC layer).
%\textbf{locally distributed excitation, inhibition and radially distributed time latency}
The proposed DPC structure selectively build the barrier against stimuli from translating and background objects that have relatively lower image angular velocities while the spatial-temporal pattern of looming objects is preferred.
Additionally, using an FFI-GD mechanism, the D-LGMD model preserves the ability to detect collision during UAV agile attitude motions including pitching, acceleration, deceleration and self-rotation. 
Systematic experiments have demonstrated that the proposed model dramatically enhances distinguish-ability of looming objects from agile-flight-derived background noise. Thus, the model is robust in handling complex dynamic visual scenes. Moreover, the proposed model functions well even with extremely small input image sizes ($38\times32$ pixels). In fact, reducing the size of input images did not harm the performance but increased the distinguish-ability towards looming objects. This notable character is likely to deliver a key success factor in energy-limited applications such as in embedded systems and MAVs.

%\normalem
\bibliographystyle{IEEEtran}
\bibliography{main}

\end{document}